\documentclass{article}

\usepackage{microtype}
\usepackage{graphicx}
\usepackage{booktabs} % for professional tables

\usepackage{hyperref}

\usepackage[accepted]{icml2024}

\usepackage{amsmath}
\usepackage{amssymb}
\usepackage{mathtools}
\usepackage{amsthm}
\usepackage{scalerel}
\usepackage{fp}
\usepackage{stackengine}
\usepackage{subfig}
\usepackage{comment}
\usepackage[capitalize,noabbrev]{cleveref}

\theoremstyle{plain}
\newtheorem{theorem}{Theorem}[section]

\newtheorem{lemma}[theorem]{Lemma}
\newtheorem{corollary}[theorem]{Corollary}
\theoremstyle{definition}
\newtheorem{definition}[theorem]{Definition}

\theoremstyle{remark}

\usepackage[textsize=tiny]{todonotes}

\DeclareMathOperator*{\ei}{Ei}

\DeclareMathOperator*{\argmax}{\arg\!\max}

\newcommand\lsb{\left[}
\newcommand\rsb{\right]}
\newcommand\abs[1]{\left|#1\right|}

\newcommand\kl[2]{\mathcal D_{\mathrm{KL}} \lsb #1 \big\| #2\rsb}

\def\clipeq{\!\mathrm{=}\!}
\def\Lla{\longleftarrow\!\!\!\!}
\def\Lra{\!\!\!\!\longrightarrow}
\newcommand\xleftrightarrows[3][0]{%
  \mathrel{%
  \ifthenelse{\equal{#1}{0}}%
  {\stackunder[2pt]{\stackon[3pt]{$\Lla\Lra$}%
     {$\scriptstyle#2$}}{$\scriptstyle#3$}}%
  {\stackunder[2pt]{\stackon[3pt]{$\Lla\hstretch{#1}{\clipeq}\Lra$}%
     {$\scriptstyle#2$}}{$\scriptstyle#3$}}%
  }%
}
\newcommand\theexpectation{\mathbb E_{\mathbf{y} \sim \mathcal{D}(\mathcal{R}, \mathbf{W}, \alpha)}}

\icmltitlerunning{Latent Optimal Paths by Gumbel Propagation for Variational Bayesian Dynamic Programming}

\begin{document}

\twocolumn[
\icmltitle{Latent Optimal Paths by Gumbel Propagation for Variational Bayesian Dynamic Programming}



\icmlsetsymbol{equal}{*}

\begin{icmlauthorlist}
\icmlauthor{Xinlei Niu}{anu}
\icmlauthor{Christian Walder}{deepmind}
\icmlauthor{Jing Zhang}{anu}
\icmlauthor{Charles Patrick Martin}{anu}
\end{icmlauthorlist}

\icmlaffiliation{anu}{Australian National University, Canberra, Australia}
\icmlaffiliation{deepmind}{Google, DeepMind}

\icmlcorrespondingauthor{Xinlei Niu}{xinlei.niu@anu.edu.au}
% \icmlcorrespondingauthor{Christian Walder}{cwalder@google.com}
% \icmlcorrespondingauthor{Jing Zhang}{xinlei.niu@anu.edu.au}

\icmlkeywords{Latent Optimal Path, Variational Bayesian Inference, Variational Autoencoder}

\vskip 0.3in
]



\printAffiliationsAndNotice{}  % leave blank if no need to mention equal contribution

\begin{abstract}
We propose the stochastic optimal path which solves the classical optimal path problem by a probability-softening solution. This unified approach transforms a wide range of DP problems into directed acyclic graphs in which all paths follow a Gibbs distribution. We show the equivalence of the Gibbs distribution to a message-passing algorithm by the properties of the Gumbel distribution and give all the ingredients required for variational Bayesian inference of a latent path, namely Bayesian dynamic programming (BDP).
We demonstrate the usage of BDP in the latent space of variational autoencoders (VAEs) and propose the BDP-VAE which captures structured sparse optimal paths as latent variables. This enables end-to-end training for generative tasks in which models rely on unobserved structural information. At last, we validate the behavior of our approach and showcase its applicability in two real-world applications: text-to-speech and singing voice synthesis. Our implementation code is available at \url{https://github.com/XinleiNIU/LatentOptimalPathsBayesianDP}.
\end{abstract}

\section{Introduction} \label{introduction}
Optimal paths are often required in many generative tasks
such as speech, music, and language modelling~\citep{kim2020glow, li2022danceformer,cai2019dtwnet}. These tasks involve the simultaneous identification of structured relationships between data and conditions. 
% Obtaining optimal paths given a graph constraint with weights to achieve an improved fit based on extracted unobserved structural features. 
Finding optimal paths given a graph constraint with learned weights can highlight unobserved structural relationships that are not immediately apparent, thus potentially improving the interpretability and effectiveness of the models.
In the context of optimal path problems, such as finding the shortest path in a graph, dynamic programming (DP) efficiently computes the solution by breaking the problem down into several sub-problems and finding optimal solutions within the sub-problems iteratively.
% The optimal paths problem can be solved by dynamic programming (DP) which breaks the problem down into several sub-problems and finds optimal solutions within the sub-problems iteratively.

% Limitation of current DP, 1. soft approximate, 2. multiple training phase

Since the DP algorithm finds shortest paths using the max operator,
it is non-differentiable which limits the usage of optimal paths in neural networks where gradient back-propagation is applied. As a workaround, previous works have approximated the max operator with smoothed functions to allow differentiation of DP algorithms~\citep{verdu1987abstract}.
However, smoothed approximations lose the sparsity of solutions which makes hard assignments become soft assignments.
Alternatively, some real-world generative applications~\citep{ren2019fastspeech, ren2020fastspeech,JeongKCCK21,li2022danceformer,halperin2019dynamic,peng2020non,liu2022diffsinger,popov2021grad}, that require to integrate structured optimal paths, split the training strategies in which neural network models depend on sparse outputs from external DP aligners~\citep{McAuliffeSM0S17,johnson2005penn} or pre-trained models~\citep{abs-1809-08895}.
However, these external components involve more than one training phase thus the model performance critically relies on them.

% In this way, I'm not emphasis why we need DP on VAEs, but emphasis we explore a new way to integrate the sparse DP on generative modelling.
In this work, we explore a novel and unified method to obtain structured sparse optimal paths with DP and showcase its application in the latent space of variational autoencoders (VAEs). Instead of a smoothed approximation for the classical optimal path problem, we propose the stochastic optimal path, which is a probabilistic softening solution by defining a Gibbs distribution where the energy function is the path score. We show this to be equivalent to a message-passing algorithm on the directed acyclic graphs (DAG) using the max and shift properties of the Gumbel distribution. 
% Then the optimal path can be sampled by a stochastic way.
% To make VAEs learn latent optimal paths, w
To learn the latent optimal paths, we give tractable closed-form ingredients for variational Bayesian inference (i.e., likelihood and KL divergence)  using DP, namely Bayesian dynamic programming (BDP), as well as an efficient sampling algorithm, which enables VAEs to obtain latent optimal paths within a DAG and achieve end-to-end training on generative tasks that rely on sparse unobserved structural relationships. We make the following contributions: 

(1) We present a unified framework that gives a probabilistic softening of the classical optimal path problem on DAGs. We notably give efficient algorithms in linear time for sampling, computing the likelihood, and computing the KL divergence, thereby providing all the ingredients required for variational Bayesian inference with a latent optimal path. 

(2) We introduce BDP-VAE framework that learns sparse optimal paths in the latent space. In the case of conditional generation, the data is not observed during inference, it is difficult to form the distribution statistics (i.e., edge weights) in the prior encoder.
% To solve the limitation of the proposed KL divergence, 
We give an alternative and flexible method to form the distribution statistics on the conditional prior by making use of a flow-based model.

(3) We demonstrate how the BDP-VAE achieves end-to-end training on two real-world challenging applications (i.e., text-to-speech (TTS) and singing voice synthesis (SVS)) and verify the behaviour of the stochastic optimal paths, latent paths and hyper-parameters proposed in \cref{sec:bayesianDP} and \cref{sec:BDP-VAE}.

\section{Related Work} \label{sec:realted}

% We present a new VAE~\cite{KingmaW13} framework learning sparse latent optimal path by DP with Gumbel propagation. 
Since traditional DP finding optimal paths is non-differentiable, there exist many alternative works that integrate DP into the neural networks by involving a convex optimization problem~\citep{AmosK17,Djolonga017}. Instead, \citet{mensch2018differentiable} proposed a unified DP framework by turning a broad class of DP problems into a DAG and obtaining the optimal path by a max operator smoothed with a strongly convex regularizer. This work can be applied in structured prediction tasks~\citep{bakir2007predicting} under supervised learning. Inspired by~\citet{mensch2018differentiable}, we proposed a probabilistic softening solution to seek stochastic optimal paths with a path distribution under a DAG. Graphical models such as Bayesian networks~\citep{Heckerman98} learn dependencies of random variables based on a DAG, our method treats the paths of a DAG, not the nodes, as random variables of a Gibbs distribution.

In many conditional generative tasks, models usually rely on structured dependencies of data and conditions, in which the dependencies are unobserved. \citet{ren2020fastspeech} and \citet{liu2022diffsinger} use a multiple training strategy by obtaining the sparse dependencies from an external DP-based techniques~\citep{McAuliffeSM0S17} at first, then use the outputs as additional inputs to the model. \citet{kim2020glow} integrates a DP on a Glow-based model to obtain unobserved monotonic alignment in parallel. Other models with structured latent representations such as HMMs~\citep{Rabiner89} and PCFGs~\citep{PetrovK07} make strong assumptions about the model structure, which could limit their flexibility and applicability. 
Instead of these works, we propose a unified framework to enable the VAEs~\citep{KingmaW13} to capture structural latent variables (i.e., sparse optimal paths), allowing for flexible adaptation to a variety of downstream tasks.

% To capture unobserved structured features such as alignment on generative models, an attention mechanism is commonly applied.
The attention mechanism~\cite{vaswani2017attention} is widely applied for obtaining unobserved dependencies in many seq-to-seq tasks.
% 1. Hard to achieve structural constraints
\citet{DengKCGR18} makes use of the properties of VAEs and captures the unobserved non-structural dependencies by learning latent alignment with attention in VAEs. However, to obtain structural constraints, attention strongly relies on model structures and other techniques such as DP to extract marginalization of the attention alignment distribution~\cite{yu2016a,yu2016neural}. Different from latent alignment with attention, we target solving the optimal path problem in a unified framework that can be easily adapted to any structural constraint by defining DAGs (e.g., structural alignment).
% our latent optimal paths with BPD can be easily adapted to any structural constraint by defining DAGs. 
By capturing unobserved sparse shortest paths under the defined DAG in a latent space, our BDP-VAE facilitates the development of more explicit structural unobserved dependencies for a variety of applications.
% 2. Computational complexity
Secondly, attention mechanisms can be computationally expensive, especially for large input sequences. Solutions such as \citet{ChiuR18} reduce the computational complexity by involving a DP. Our method seeks stochastic optimal paths that occur in linear time with respect to edge numbers of DAGs (\cref{cor:linear_time}).

The Gumbel-Max trick makes use of the max property of the Gumbel distribution which allows for efficient sampling from discrete distributions~\citep{maddison2014sampling}. \citet{JangGP17} and \citet{MaddisonMT17} facilitate gradient-based learning for Gibbs distribution by relaxing the component-wise optimization in the Gumbel-Max trick.
\citet{StruminskyGRKV21} focuses on leveraging the Gumbel-Max trick on the score function estimator. Unlike those, our research leverages the max and shift properties of Gumbel distribution for message-passing on DP to obtain ingredients required for variational Bayesian inference for latent paths.

\section{Preliminaries} \label{sec:background}
% \section{MATHEMATICAL PRELIMINARIES}
This section provides background on the notation definition of a DAG, a definition of the traditional optimal path problem given a DAG, and properties of the Gumbel random variable.

\noindent\textbf{Definition of a Graph:}
We denote $\mathcal{R} = (\mathcal{V},\mathcal{E})$ be a directed acyclic graph with nodes $\mathcal{V}$ and edges $\mathcal{E}$. Assume without loss of generality that the nodes are numbered in topological order, such that $\mathcal V=(1,2,\dots,N)$ and $u<v$ for all $(u,v)\in \mathcal E$. Further, we assume that $1$ is the only node without parents and $N$ the only node without children. We denote the edge weights $\mathbf W\in\mathbb R^{N\times N}$ with $ w_{i,j}=-\infty$ for all  $(u,v)\not\in\mathcal E$. Let $\mathcal Y(1,v)$ be the set of all paths from $1$ to $v$. Associate with each path $\mathbf y=(y_1,y_2,\dots,y_{|{\mathbf y}|})$ a score obtained by summing edge weights along the path, 
% so that
defined as $ || \mathbf y ||_{\mathbf W}=\sum_{i=2}^{|{\mathbf y}|} w_{y_{i-1},y_i}=\sum_{(u,v)\in \mathbf y} w_{u,v}$, where the final expression introduces the notation $(u,v)\in \mathbf y$ for the edges $(u,v)$ that make up path $\mathbf y$. Denote the set of parents of node $v$ by $\mathcal P(v)=\{u: (u,v)\in\mathcal E\}$ and the set of children of node $u$ by $\mathcal C(u)=\{v: (u,v)\in\mathcal E\}$.

\noindent\textbf{Non-Stochastic Optimal Paths:}
The traditional optimal path problem is to find the highest scoring path from node $1$ to node $N$,
\begin{align}
	\mathbf y^* = \argmax_{\mathbf y\in\mathcal Y(1,N)} || \mathbf y ||_{\mathbf W}.
\end{align}
This can be solved in $O(\abs{\mathcal E})$ time by iterating in topological order. The score $\xi(\cdot)$ is defined as
\begin{align}
	\xi(1) & = 0 \\
	\forall v\in\left\{2,3,\dots,N\right\}, ~~~ \xi(v) & = \max_{u\in\mathcal P(v)} \xi(u) + w_{u,v},
\end{align}
after which $\mathbf y^*$ is obtained (in reverse) by tracing from $N$ to $1$, following the path of nodes $u$ for which the maximum was obtained in the above.

\noindent\textbf{Gumbel Random Variable:}
Let $\mathcal G(\mu)$ denote the unit scale Gumbel random variable with location parameter $\mu$ and probability density function 
\begin{align}
	\mathcal G(x|\mu) = \exp(-(x-\mu)-\exp(x-\mu)).
\end{align}
We now review the properties of the Gumbel which we will exploit in~\cref{sec:bayesianDP}.
% the sequel.
Let $X\sim\mathcal G(\mu)$. The Gumbel distribution is closed under shifting, with
\begin{align}
	X+\mathrm{const.}&\sim\mathcal G(\mu+\mathrm{const.}).
	\label{eqn:gumbelshift}
\end{align}
Let $X_i\sim\mathcal G(\mu_i)$ for all $i \in \left\{1,2,\dots,m\right\}$.
The Gumbel is also closed under the $\max$ operation, with
\begin{align}
	\max(\left\{X_1,X_2,\dots,X_m\right\}) \sim\mathcal G(\log\sum_{i=1}^m \exp(\mu_i)).
	\label{eqn:gumbelmax}	
\end{align}
Finally, there is a closed-form expression for the index which obtains the maximum in the above expression:
\begin{align}
	p(k = \hspace{-2mm}\argmax_{i \in \{1,2,\dots,m\}}\hspace{-1mm}X_i)
	=
	\frac{\exp(\mu_k)}{\sum_{i=1}^m \exp(\mu_i)}.
	\label{eqn:gumbelargmax}
\end{align}

\section{Bayesian Dynamic Programming
% \protect\footnote{Proofs for each lemma in this section are provided in the supplementary material}
} \label{sec:bayesianDP}

In this section, we propose a stochastic approach to seek optimal paths. We denote a distribution family given a DAG with edge weights and give ingredients required for variational Bayesian inference by using DP with Gumbel propagation, namely, Bayesian dynamic programming.

\subsection{Stochastic Optimal Paths}

In the stochastic approach, every possible path $\mathbf{y} \in \mathcal{Y}$ on $\mathbf{W}$ follows a Gibbs distribution given a DAG $\mathcal{R}$, edge weights $\mathbf{W}$ and temperature parameter $\alpha$ defined by \cref{def:gibbs}. 

\begin{definition} \label{def:gibbs}
Denote by
\begin{align}
\mathcal D(\mathcal R, \mathbf W, \alpha)
\label{eqn:rvname}	
\end{align}
the Gibbs distribution over $\mathbf y\in\mathcal Y(1,N)$ with probability mass function 
\begin{align}
	\mathcal D(\mathbf y|\mathcal R, \mathbf W, \alpha)
	& =
	\frac{\exp(\alpha \left \| \mathbf y \right \|_{\mathbf W})}
	{\sum_{\widehat {\mathbf y}\in\mathcal Y(1,N)}\exp(\alpha \left \| \widehat {\mathbf y} \right \|_{\mathbf W})}.
	\label{eqn:gibbsdistribution}
\end{align}
\end{definition}

Despite the intractable form of the denominator in \cref{eqn:gibbsdistribution}, we provide the ingredients necessary for approximate Bayesian inference for latent distribution (unobserved) $\mathcal D$. 
In particular, we can efficiently compute the normalized likelihood (\cref{cor:pathchainrule}), sample (\cref{cor:sample}), and compute the KL divergence within $\mathcal{D}(\mathcal{R},\cdot,\alpha)$ (\cref{lem:kl}) in linear time (\cref{cor:linear_time}).

\subsection{Gumbel Propagation}

The Gumbel propagation offers an equivalent formulation of \cref{def:gibbs} that lends itself to dynamic programming by the properties in \cref{eqn:gumbelargmax} and \cref{eqn:gumbelshift} as per the following result. Proofs per lemma of this subsection are in \cref{A:gumbel_proof}.

\begin{lemma}
Let 
\begin{align}
\label{eqn:y}
	Y=\argmax_{\mathbf y\in\mathcal Y(1,N)} \left\{\Omega_{\mathbf y}\right\},
\end{align}
where for all $\mathbf y \in \mathcal Y(1,N)$,
\begin{align}
	 \Omega_{\mathbf y} & = \mathcal \alpha \left\|\mathbf y\right\|_{\mathbf W}+G_{\mathbf y} \\
	 G_{\mathbf y} & \sim \mathcal G(0).
\end{align}
Then the probability of $Y=\mathbf y$ is given by \eqref{eqn:gibbsdistribution}.
\end{lemma}

Let the definitions of $\Omega_{\mathbf y}$ and $G_{\mathbf y}$ extend to all $\mathbf y \in \bigcup_{u=1}^N \mathcal Y(1,u)$, which is the set of all partial paths. We define for each node $v\in\mathcal V$ the real-valued random variable
\begin{align}
	Q_{v} 
	& = \max_{\mathbf y\in\mathcal Y(1,v)} \left\{\Omega_{\mathbf y}\right\}.
\end{align}

\begin{lemma}
The $Q_{v}$ are Gumbel distributed with
\begin{align}
	Q_{v} \sim \mathcal G(\mu_v),
\end{align}
where
\begin{align}
\label{eqn:mudefinition}
	\mu_v = \log \sum_{\mathbf y\in\mathcal Y(1,v)} \exp(\alpha\left\|\mathbf y\right\|_{\mathbf W}).
\end{align}	
\end{lemma}

The motivation for constructing $Q_{v}$ is $Q_{v}$ allows us to set up a recursion on the entire DAG $\mathcal{R}$.
We now state the first main result in \cref{lem:murecursion}.

\begin{lemma}
\label{lem:murecursion}
The location parameters $\mu_v$ satisfy the recursion
\begin{align}
	\mu_1 & = 0 \\
	\mu_v & = \log \sum_{u \in \mathcal P(v)} \exp(\mu_u + \alpha \, w_{u,v}).
	\label{eqn:murecursion}
\end{align}
for all $v\in\left\{2,3,\dots,N\right\}$.
\end{lemma}

\subsection{Sampling and Likelihood}

Then we state an alternative normalized likelihood of a sampled path $\mathbf{y}$ by~\cref{cor:sample} as~\cref{cor:pathchainrule} according to a transition matrix defined in~\cref{lem:transition}. The transition matrix in \cref{lem:transition} can be computed according to the location parameter $\mu$ defined in \cref{lem:murecursion} directly. Proofs per lemma of this subsection are in \cref{B:sampling_likelihood_proof}.

\begin{lemma}\label{lem:transition}
	Let paths $\mathbf y=(y_1,y_2,\dots,y_{\abs{\mathbf y}})$ denote the component of the random variable $Y$ defined in \eqref{eqn:y}, given that $Y =(y_1,y_2,\dots,y_{\abs{\mathbf y}})$. The probability of the transition $v \rightarrow u$ is
 % Let $\mathbf y=(y_1,y_2,\dots,y_{\abs{\mathbf y}})\sim Y$, where the random variable $Y$ is defined in \eqref{eqn:y}. The probability of the transition $v \rightarrow u$ is
	\begin{align}
	\label{eqn:transitionconditional}
		\pi_{u, v} & \equiv 
		p(y_{i-1}=u|y_{i}=v,u\in\mathcal P(v)) \\
		& = \frac{\exp(\mu_u+\alpha \, w_{u,v})}{\exp(\mu_v)},
		\label{eqn:transitionmu}
	\end{align}
	for all $i\in\{2,3,\dots,N\}$.
\end{lemma}
\begin{corollary} 
	The path probability may be written 
	\begin{align}
		\mathcal D(\mathbf y|\mathcal R, \mathbf W, \alpha) = 
	\prod_{(u,v)\in\mathbf y} \pi_{u,v}.
	\label{eqn:pathchainrule}
	\end{align}
	\label{cor:pathchainrule}
\end{corollary}
\begin{corollary}
	Paths $\mathbf y\sim\mathcal D(\mathcal R, \mathbf W, \alpha)$ may be sampled (in reverse) by 
	\begin{enumerate}
		\item Initialising $v=N$, 
		\item sampling $u\in\mathcal P(v)$ with probability $\pi_{u,v}$,
		\item setting $v\leftarrow u$,
		\item if $v = 1$ then stop, otherwise return to step 2.
	\end{enumerate}
	\label{cor:sample}
\end{corollary}

\subsection{KL Divergence}
Given $\mathcal{D}(\mathcal{R}, \mathbf{W}, \alpha)$ and $\mathcal{D}(\mathcal{R}, \mathbf{W}^{(r)}, \alpha)$, where $\mathbf{W}^{(r)}$ are edge weights have different values to $ \mathbf{W}$. We give a tractable closed-form of the KL divergence within the distribution family of $\mathcal{D}(\mathcal{R}, \cdot, \alpha)$ in \cref{lem:kl}. Proofs per lemma of this subsection are in \cref{C:KL_proof}.

\begin{definition} \label{def:omegadefinition}
We denote the total probability of paths that include a given edge $(u,v)\in\mathcal E$ by
\begin{align}
	\omega_{u,v} \equiv 
	\sum_{\{\mathbf y \in\mathcal Y(1,N): (u,v)\in \mathbf y\}}\mathcal D(\mathbf y|\mathcal R, \mathbf W, \alpha).
	\label{eqn:omegadefinition}
\end{align}
\end{definition}
The quantity in the above definition may be computed using two dynamic programming passes, one topologically ordered and the other reverse topologically ordered, by applying the following

\begin{lemma} \label{lem:omega}

For all $(u,v)\in\mathcal E$, 
\begin{align}
	\omega_{u,v} = \pi_{u,v} \, \lambda_u \, \rho_v,
\end{align}
where we have the recursions
\begin{align}
\lambda_1 & = 1 \\
\lambda_v
& = 
\sum_{u \in \mathcal P(v)} \lambda_u \, \pi_{u,v}
	\label{eqn:lambdarecursion}
\end{align}
for all $v\in\left\{2,3,\dots,N\right\}$ (in topological order w.r.t. $\mathcal R$), and
\begin{align}
        \rho_N & = 1\\
        \rho_u & = \sum_{v \in \mathcal{C}(u)} \rho_v \pi_{u,v}
\end{align}
for all $u\in\left\{N-1,N-2,\dots,1\right\}$ (in reverse topological order w.r.t. $\mathcal R$).
\end{lemma}
\begin{lemma}
The KL divergence within the family $\mathcal D(\mathcal R, \cdot, \alpha)$ is
\begin{align}
	& \kl{\mathcal{D}(\mathcal{R}, \mathbf{W}, \alpha)}{\mathcal{D}(\mathcal{R}, \mathbf{W}^{(r)}, \alpha)} 
	\nonumber
	\\ & ~~~~~~ = \mu_N^{(r)}-\mu_N + \alpha \sum_{(u,v)\in\mathcal{E}} \omega_{u,v} \big( w_{u,v}- w_{u,v}^{(r)}\big),
	\label{eqn:kld}
\end{align}
where $\omega_{u,v}$ is the marginal probability of edge $(u,v)$ on $\mathcal{D}(\mathcal{R},\mathbf{W},\alpha)$ defined in \cref{def:omegadefinition}, $\mu_N$ is defined in \cref{eqn:mudefinition} and $\mu_N^{(r)}$ is similar to $\mu_N$ but defined in terms of $\mathbf{W}^{(r)}$ rather than $\mathbf{W}$.
\label{lem:kl}
\end{lemma}

\begin{corollary} \label{cor:linear_time}
 The KL divergence \eqref{eqn:kld}, the likelihood \eqref{eqn:pathchainrule}, and the sampling algorithm (\cref{cor:sample}) may be computed in $O(\abs{\mathcal E})$ time.
\end{corollary}

\begin{figure}[t]
% \vskip -0.1in
\begin{center}
\centerline{\includegraphics[width=0.99\columnwidth]{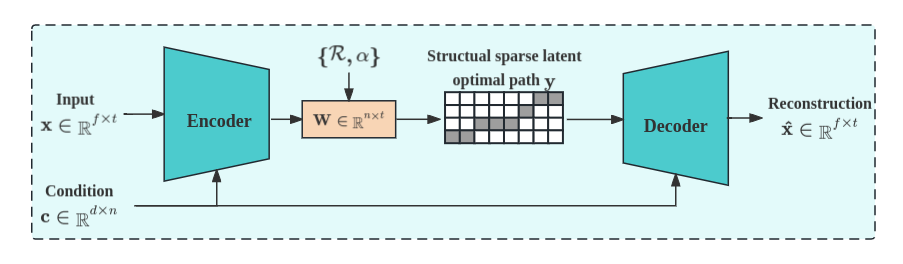}}
\caption{A pipeline of BDP-VAE. BDP-VAE captures the unobserved sparse structural dependency (i.e., optimal paths on a DAG) in the latent space in parallel training the model and allows gradient-based optimization for learning the edge weights $\mathbf{W}$.}
\label{fig.pipeline_BDPVAE}
\end{center}
\vskip -0.2in
\end{figure}

\section{BDP-VAE} \label{sec:BDP-VAE}
We now show how to apply the method in \cref{sec:bayesianDP} to a conditional VAE framework to obtain sparse latent optimal paths. An unconditional BDP-VAE framework can be directly applied based on \cref{cor:pathchainrule}, \cref{cor:sample}, and \cref{lem:kl}, but a conditional BDP-VAE framework may be challenging in real applications. Given a sequential-like input $\mathbf{x}$ with length $t$ and a sequential-like condition $\mathbf{c}$ with length $n$, where $t\neq n$ and $\{t,n\}$ are varying within the dataset. We wish to find an unobserved hard structural relationship between $\mathbf{x}$ and $\mathbf{c}$ in the latent space of VAEs denoted as $\mathbf{y}$. Conditional BDP-VAEs consist of three parts: an encoder models posterior distribution $q(\mathbf{y}|\mathbf{x},\mathbf{c}; \phi)$, a decoder models the distribution of $p(\mathbf{x}|\mathbf{y},\mathbf{c}; \theta)$, and a prior encoder models the prior distribution $p(\mathbf{y}|\mathbf{c};\theta)$. An overview pipeline of BDP-VAE is in \cref{fig.pipeline_BDPVAE}, which captures structured sparse optimal paths in the latent space.

We assume the conditional input $\mathbf c$ is always observed and the conditional ELBO is defined as 
\begin{equation}
\begin{split}
	\mathcal L(\phi, \theta, \mathbf x | \mathbf c)
	= 
	\mathbb E_{\mathbf y \sim q(\cdot|\mathbf x, \mathbf c; \phi)} 
	\left[
		\log p(\mathbf x|\mathbf y, \mathbf c; \theta)
	\right] \\
	- 
	\kl{q(\mathbf y|\mathbf x, \mathbf c; \phi)}{p(\mathbf y|\mathbf c; \theta)}.
\end{split}
\label{eqn:elbo}
\end{equation}

\subsection{Posterior and Latent Optimal Paths}

Given a DAG $\mathcal{R}$ with edge $\mathcal{E}$ and nodes $\mathcal{V}$, the distribution of the posterior encoder is denoted as 
\begin{equation}
    q(\mathbf{y}|\mathbf{x},\mathbf{c}; \phi) = \mathcal{D}(\mathbf{y}| \mathcal{R}, \mathbf{W} = \text{NN}_\mathbf{W}(\mathbf{x},\mathbf{c};\phi),\alpha)
\end{equation}
where NN$_\mathbf{W}(\cdot;\phi)$ is a neural network to learn the edge weights $\mathbf{W}$ of the DAG $\mathcal{R}$ and $\alpha$ is a hyper-parameter. The latent optimal path $\mathbf{y}$ with $\{0,1\}$ can be sampled reversely according to \cref{lem:transition} and \cref{cor:sample}. As a result, $\mathbf{y}$ forms a \textit{sparse matrix} with dimensions identical to those of the weight matrix $\mathbf{W}$. As shown in \cref{fig.pipeline_BDPVAE}, the sparsity and structure of $\mathbf{y}$ are inherently achieved through its construction using the DAG $\mathcal{R}$\footnote{In the case where the structure of DAGs $\mathcal{R}$ is unknown. Since we are learning the DAG weights $\mathbf{W}$ and zero weights are equivalent to removing an edge, in some sense the BDP algorithm can in principle learn an approximate DAG structure by imposing small edge weight values. However, in this study, we target to verify our method on applications with a clear prior knowledge of structure DAGs.}. 

% Thus, $\mathbf{y}$ is a \textit{sparse matrix} with the same size as the weight matrix $\mathbf{W}$ that achieves sparsity and structure by construction with the DAG $\mathcal{R}$ (see pseudo \cref{alg:DTW_backward} and \cref{alg:MA_backward} of computational examples in \cref{sec:DTW} and \cref{sec:MA}). 

% In the case where the structure of DAGs $\mathcal{R}$ is unknown, since we are learning the DAG weights $\mathbf{W}$ and zero weights are equivalent to removing an edge, in some sense the BDP algorithm can in principle learn an approximate DAG structure by imposing small edge weight values. However, in this study, we target to verify our method on applications with a clear prior knowledge of structure DAGs.

\subsection{Conditional Prior}\label{sec:condition_prior}

Denote the distribution of the conditional prior as
\begin{equation}
    p(\mathbf{y}|\mathbf{c}; \theta) = \mathcal{D}(\mathbf{y}| \mathcal{R}, \mathbf{W}^{(0)}= \text{NN}_\mathbf{W^{(0)}}(\mathbf{c};\theta), \alpha)
\end{equation}
We have provided a closed-form KL divergence within the distribution family $\mathcal D(\mathcal R, \cdot, \alpha)$ in \cref{lem:kl} which may be convenient for unconditional generation by pre-setting the prior distribution statistics directly or has tractable $\mathbf{W}^{(0)}$.
 
In most conditional generation tasks, the non-accessible $\mathbf{x}$ during the inference phase in real applications leads to problems on the prior encoder when forming the edge weights $\mathbf{W}^{(0)}$ given information on $\mathbf{c}$ only, especially in the case that $\mathbf{x}$ has varying lengths $t$. 
% This leads to an intractable $\mathbf{W}^{(0)}$ during the inference phase.
To address this issue,
% it, 
we give a flexible solution for inferring feature information of $\mathbf{x}$ given $\mathbf{c}$ to form
the edge weights $\mathbf{W}^{(0)}$ in the conditional prior.
Inspired by \citet{flowseq}, we make use of a flow-based model~\footnote{The flow-based conditional prior strategy is proposed to solve the problem of forming edge weights $\mathbf{W}^{(0)}$ in case of $\mathbf{W}^{(0)}$ is intractable if $\mathbf{x}$ is non-accessible with varying lengths $t$. In practical applications, this strategy is not mandatory if the task has a known and fix prior $\mathcal{D}(\mathcal{R},\mathbf{W}^0,\alpha)$ or tractable $\mathbf{W}^{(0)}$ (i.e., \cref{lem:kl}).} as the conditional prior to infer information about $\mathbf{x}$ condition on $\mathbf{c}$ and further obtain edge weights $\mathbf{W}^{(0)}$.
 
 % The feature information of $\mathbf{x}$ can be approximated by a normalized flow given the conditional inputs of $\mathbf{c}$. 
Assume there exists a series of invertible transformations of random variables $\mathbf{x}$, such that 
\begin{equation}
    \mathbf x  \xleftrightarrows{f_1}{g_1} \cdots  \xleftrightarrows{f_k}{g_k} \mathbf c \xleftrightarrows{f_{k+1}}{g_{k+1}} \cdots\xleftrightarrows{f_K}{g_K} v
\end{equation}
where $f = f_1 \circ \cdots \circ f_K$ and $v\sim N(0,1)$ ($\theta$ is omitted for brevity),
% We learn a complicated distribution $p_{\mathbf{x}|\mathbf{c}}(\mathbf{x}|\mathbf{c})$ using a the prior $p(\mathbf{c})$ and a
% mapping $f_{\theta}: \mathbf{x}\times \mathbf{v} \rightarrow \mathbf{c}$, which is bijective in $\mathbf{x}$ and $\mathbf{c}$.
the KL divergence term in \cref{eqn:elbo} can be written as
\begin{align}
	& \kl{q(\mathbf y|\mathbf x, \mathbf c; \phi)}{p(\mathbf y|\mathbf c; \theta)} 
	\\
	& ~~ = 
	\kl{q(\mathbf y|\mathbf x, \mathbf c; \phi)}{p(\mathbf y|\mathbf x, \mathbf c; \theta)}
	- \log p(\mathbf x|\mathbf c; \theta)  \nonumber\\
	& ~~ = 
	- \log p_{N(0,1)}(f_\theta(\mathbf x))|\text{det}(\frac{\partial f_\theta(\mathbf x)}{\partial \mathbf x})|
\end{align}
% \begin{align}
% 	  \kl{q(\mathbf y|\mathbf x, \mathbf c; \phi)}{p(\mathbf y|\mathbf c; \theta)}  & = 
% 	% \\
% 	% & ~~ = 
% 	\kl{q(\mathbf y|\mathbf x, \mathbf c; \phi)}{p(\mathbf y|\mathbf x, \mathbf c; \theta)}
% 	- \log p(\mathbf x|\mathbf c; \theta) \\
% 	& ~~ = 
% 	- \log p_{N(0,1)}(f_\theta(\mathbf x))|\text{det}(\frac{\partial f_\theta(\mathbf x)}{\partial \mathbf x})|
% \end{align}
The backward pass is to infer the KL divergence during training. The forward pass is to infer feature information of $\mathbf x$ given $\mathbf c$ and form edge weight $\mathbf W^{(0)}$ during inference.

\subsection{Learning} \label{sec:learning}

Based on the idea of \citet{mohamed2020monte}, the gradient of the ELBO \eqref{eqn:elbo} with respect to $\theta$ is straightforward, however, the gradient with respect to $\phi$ of the reconstruction error part in the ELBO is non-trivial. We make use of the REINFORCE estimator
%%  Add proof to appendix, gap of \approx
\begin{align} \label{eqn:policygradient}
	& \nabla_\phi \,
	\mathbb E_{\mathbf y \sim q(\cdot|\mathbf x, \mathbf c; \phi)}
	\left[
		\log p(\mathbf x|\mathbf y, \mathbf c; \theta)
	\right]
	\\
	& 
	~~~~~~ =
			\log p(\mathbf x|\mathbf {\tilde y}, \mathbf c; \theta) 
		\nabla_\phi \, \log q(\mathbf {\tilde y}|\mathbf x, \mathbf c; \phi),
	\nonumber
\end{align}
where $\mathbf {\tilde y}$ is an exact sample via \cref{cor:sample} from the posterior $q(\cdot|\mathbf x, \mathbf c; \phi)$, and we recall that $\log q(\mathbf {\tilde y}|\mathbf x, \mathbf c; \phi)$ may be computed using the efficient and exact closed-form of \cref{eqn:pathchainrule}, and automatically differentiated. 

Alternatively, we provide hints of Gumbel softmax trick to avoid using the REINFORCE estimator in \cref{sec:gumbelsoftmaxtrick} for interested readers. In this study, we focus on evaluating the closed form of latent sparse optimal paths. Compared to the reparameterization trick discussed in \cref{sec:gumbelsoftmaxtrick}, the log-derivative method (i.e., the REINFORCE estimator) has the advantage of being the most straight-forward, is formally unbiased, does not require the temperature parameter, and allows fast evaluation with the closed form expression of sparse paths $\mathbf{y}$ in \cref{eqn:pathchainrule}. 
% In this study, our motivation is to maintain the sparsity of optimal paths.

\section{Experiments} \label{sec:experiments}

\begin{figure*}[t]
% \vskip 0.2in
\begin{center}
\centerline{\includegraphics[width=\textwidth]{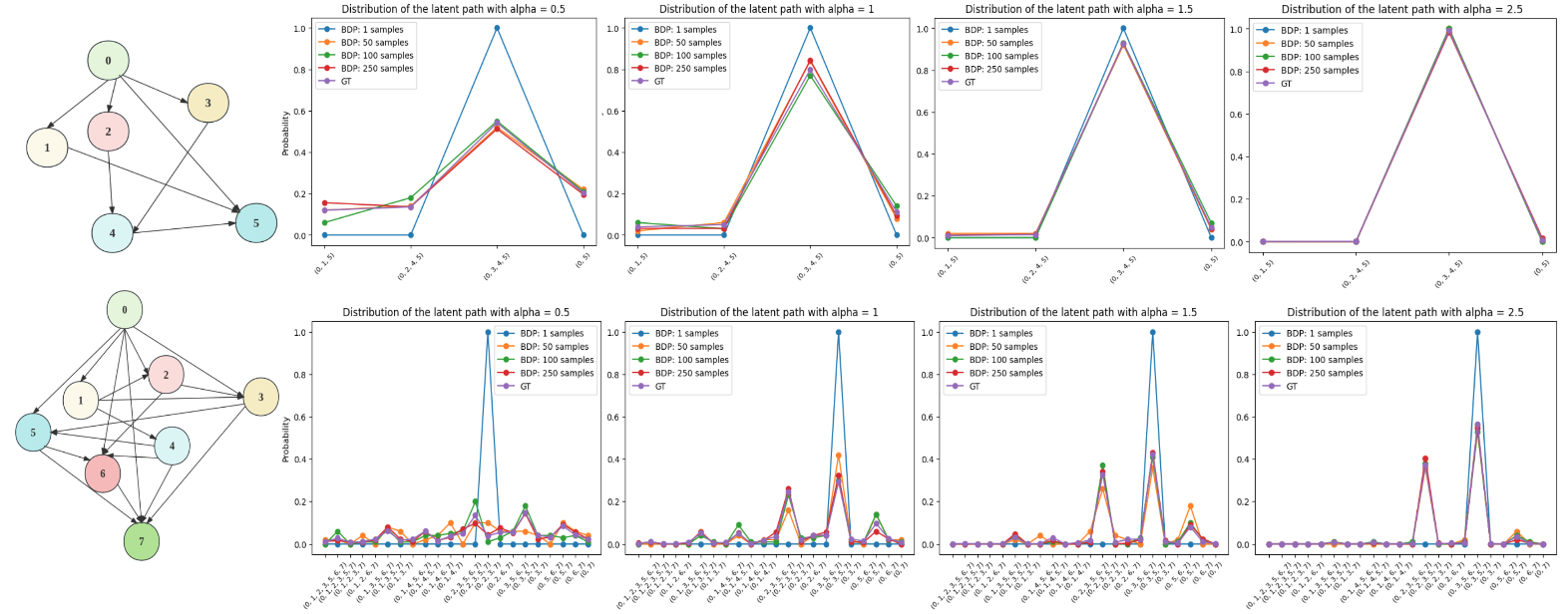}}
\caption{ Toy experiments on BDP to find stochastic optimal paths under randomly generated DAGs. \textit{The first row is a 5-node DAG and its density plots with different $\alpha$ value. The second row is an 8-node DAG and its density plots with different $\alpha$ value.}}
\label{fig.toy_exp}
\end{center}
\vskip -0.2in
\end{figure*}

In this section, we conduct four experiments to verify our methods from \cref{sec:bayesianDP} and \cref{sec:BDP-VAE}, and show its applicability on two real-world applications.
To show the generalization of methods in \cref{sec:bayesianDP}, we extend BDP to two common application examples of computational graphs: monotonic alignment (MA)~\citep{kim2020glow} and dynamic time warping (DTW) ~\citep{sakoe1978dynamic, mensch2018differentiable}. Details of examples of computational graphs, pseudo-codes, and time complexity analysis are provided in \cref{D:example}. We provide detailed model architecture used in our experiments in \cref{E:model} and experimental details and setup in \cref{EE:exp}.
% Model architectures and experimental details are provided in \cref{E:model} and \cref{EE:exp}.
% the model architecture, experimental details, examples of computational graphs and pseudo-codes are provided in \cref{D:example}, \cref{E:model}, and \cref{EE:exp}.
% \textit{Details of the model architecture, experiment details, example computational graphs, and pseudo-code are provided in the supplementary material}.
% \footnote{Details of the example computational graphs and pseudo-code are provided in the supplementary material}. 

\subsection{Stochastic Optimal Paths on Toy DAGs}\label{sec:toy}

We conduct an experiment to demonstrate how BDP in \cref{sec:bayesianDP} finds the optimal paths on toy DAGs, in which the DAG structures and edge values are randomly generated. In \cref{fig.toy_exp}, we plot approximated density plots of path distributions with 1, 50, 100, and 250 samples by BDP compared with the corresponding ground truth path density plot. As BDP samples increase, the distribution of path samples will eventually converge to the real path distribution.

We then studied how the value of the temperature parameter $\alpha$ affects the path distribution. The larger $\alpha$ is, the sharper the distribution is, leading to an accurate optimal path result with less sample time. Conversely, for too small $\alpha$, the BDP needs more samples to obtain the optimal path.
We conduct another experiment to connect this finding with BDP-VAE in \cref{F:discussion}.

\begin{table*} [t]
  \renewcommand{\tabcolsep}{1.5mm}
  \caption{Mel Cepstral Distortion (MCD) and Real-Time Factor (RTF) compared with other TTS models.}
  \footnotesize
  % \vskip 0.05in
  \label{tab:MCD}
  \centering
  \begin{tabular}{llllll}
    \toprule
    \textbf{Model} &  \textbf{Training} &  \textbf{Align.(Train)} & \textbf{Align. (Infer)} & \textbf{MCD} & \textbf{RTF} \\
    \midrule
    FastSpeech2 \citep{ren2020fastspeech} (Baseline) & Non end-to-end & Discrete & Continuous & 9.96 $\pm$ 1.01 & 3.87 $\times 10^{-4}$ \\
    Tacotron2 \citep{tacotron2} & End-to-end & Continuous & Continuous & 11.39 $\pm$ 1.95 & 6.07 $\times 10^{-4}$ \\ 
    VAENAR-TTS \citep{lu2021vaenar}& End-to-end & Continuous & Continuous & 8.18 $\pm$ 0.87 &1.10 $\times 10^{-4}$\\
    Glow-TTS \citep{kim2020glow} & End-to-end & Discrete & Continuous & 8.58 $\pm$ 0.89 & 2.87 $\times 10^{-4}$\\ 
    % BVAE-TTS \cite{lee2021bidirectional} & - & -\\ 
    BDPVAE-TTS (ours) & End-to-end & Discrete & Discrete  & 8.49 $\pm$ 0.96 & 3.00  $\times 10^{-4}$\\
    \bottomrule
  \end{tabular}
  % \vskip -0.15in
\end{table*}

\begin{figure}[ht]
% \vskip 0.2in
\begin{center}
% \centerline{\includegraphics[width=\columnwidth]{Figure/Exp_2_spect.png}}
\centerline{\includegraphics[width=0.5\textwidth]{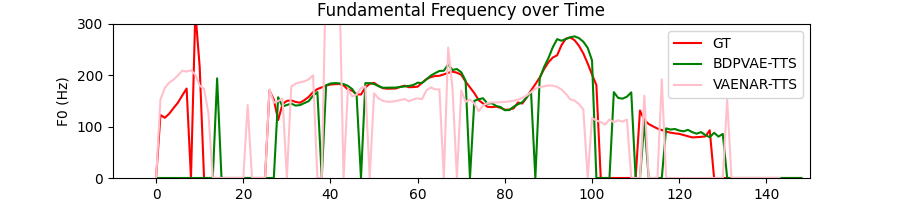}}
\caption{Inference F0 trajectory comparison with VAENAR-TTS of utterance "I suppose I have many thoughts.". \textit{The intonation of BDPVAE-TTS is close to the GT indicating that sparse optimal paths help the decoder with a better understanding of how phoneme contributes to the overall utterance with approximated durations.}}
\label{fig.Exp2_f0}
\end{center}
\vskip -0.2in
\end{figure}

\subsection{Application: End-to-end Text-to-Speech}\label{sec:tts}

\begin{figure}[t]
% \vskip 0.2in
\begin{center}
\centerline{\includegraphics[width=0.95\columnwidth] {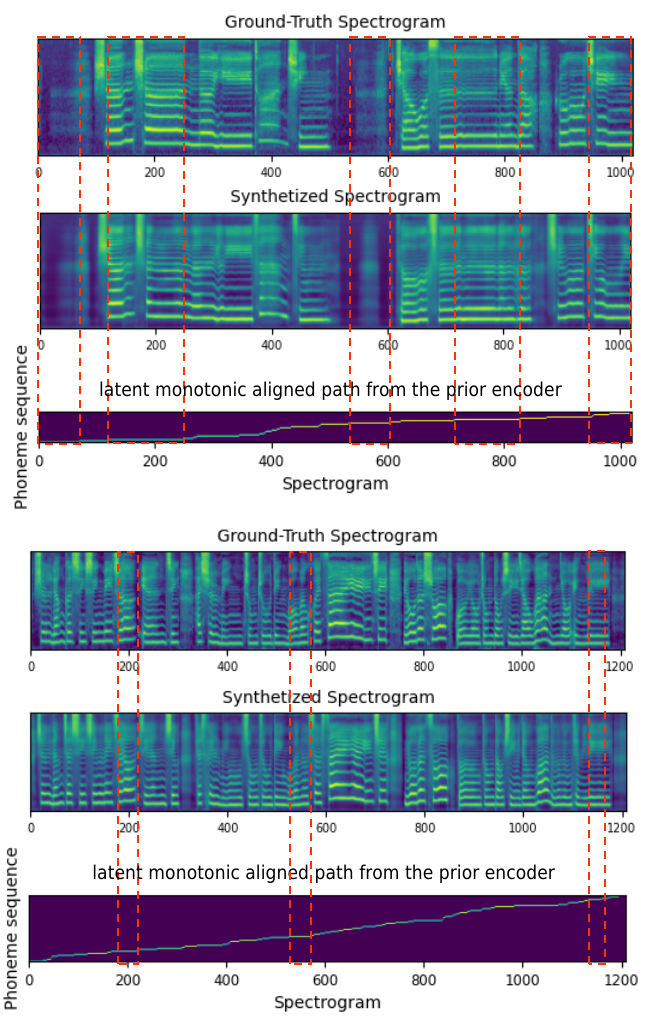}}
\caption{Visualization of GT, synthesized singing voice spectrogram, and latent optimal path from the prior encoder. \textit{The GT and generated spectrogram are almost identical, and the generated spectrogram has a similar temporal structure to the inferred latent optimal path.}}
\label{fig.Exp3_spect}
\end{center}
\vskip -0.3in
\end{figure}

We apply the BDP-VAE framework with the computational graph of MA to perform an end-to-end TTS model on the RyanSpeech~\citep{zandie2021ryanspeech} dataset.  RyanSpeech contains 11279 audio clips (10 hours) of a professional male voice actor’s speech recorded at 44.1kHz. We randomly split 2000 clips for validation and 9297 clips for training.

The task of TTS involves an unobserved monotonic hard alignment that maps phonemes to time intervals, since the order of the phonemes should be preserved while the duration of each may vary in speech. Thus, TTS models generate speech according to corresponding phonemes in which the duration of each phoneme is discrete, structural, and unobserved~\cite{mehrish2023review}.
To show the BDP-VAE framework can be easily adapted into down-stream tasks, we redesigned the popular non-end-to-end TTS model FastSpeech2~\citep{ren2020fastspeech} into the BDP-VAE framework, namely BDPVAE-TTS, which can capture unobserved hard monotonic dependencies between phonemes and utterances jointly on both training and inference.

We verify model performance by an objective metric, the Mel cepstral distortion (MCD)~\citep{kubichek1993mel,ChenTQZLW22}, between ground truths and synthesized outputs. We record inference speed by real-time factor (RTF) per generated spectrogram frame. We randomly pick 70 sentences from the test set, the numerical results are shown in \cref{tab:MCD}. Our method outperforms the baseline on both MCD and RTF and achieves end-to-end training. This shows the success of BDP-VAE in adapting a non-end-to-end model that relies on external DP aligners into an end-to-end pipeline.

We make a comparison with other end-to-end TTS models~\cite{tacotron2,kim2020glow,lu2021vaenar} which target capturing phoneme dependencies. \citet{tacotron2} models the unobserved temporal dependencies by an auto-regressive architecture with attention. \citet{kim2020glow} integrate a monotonic alignment search in parallel in a Glow model~\cite{kingma2018glow} to obtain hard monotonic alignment.  \citet{lu2021vaenar} utilizes the latent Gaussian and captures soft monotonic alignment in the decoder by causality-masked self-attention. 
Among them, BDPVAE-TTS performs discrete monotonic alignment on both training and inference that ensures model consistency for training and inference. BDPVAE-TTS gets a better MCD and RTF than~\citet{tacotron2} and~\citet{kim2020glow}, but gets higher MCD and RTF than~\citet{lu2021vaenar}.

For the RTF, since BDPVAE-TTS involves a linear time consumption DP algorithm which causes higher inference time than VAENAR-TTS. We have discussed time complexity for capturing phoneme-to-frame dependencies during training for each end-to-end TTS model in this experiment as below. However, the linear time consumption (\cref{cor:linear_time}) is the best we can do for solving a DP problem. 

\noindent\textbf{Time Complexity Analysis:} Denote the maximum number of spectrogram frames in the computational graph of MA is $T_{mel}$ and the maximum number of phoneme tokens in the computational graph of MA is $T_{text}$.
% For the RTF, t
In this experiment, BPDVAE-TTS obtains discrete latent monotonic paths by BDP with $\mathcal{O}(T_{mel})$ time complexity (detail implementations and discussion could be found in \cref{E:MA_time}). Besides, the time complexity of monotonic alignment search (MAS) in Glow-TT is $\mathcal{O}(T_{text}\times T_{mel})$~\cite{kim2020glow}, the time complexity of self-attention in VAENAR-TTS is $\mathcal{O}(1)$~\cite{vaswani2017attention}, and the time complexity of auto-regressive TTS model (i.e., Tactron2~\cite{tacotron2}) is $\mathcal{O}(T_{mel})$.

For the MCD, even though BDPVAE-TTS obtains a higher MCD than VAENAR-TTS, BDPVAE-TTS captures \textit{discrete monotonic aligned paths} in its latent space on training and inference phase which ensures model's train and test consistency and improves the model's interpretability.
VAENAR-TTS compresses the learned feature of spectrograms with conditions into Gaussian latent variables and uses these in its decoder for reconstruction. During the decoding process, these latent variables are used alongside phonemes to reconstruct the outputs.
% The decoder not only reconstructs information from conditions but also the latent distribution of inputs. 
In contrast, the latent variables in BDPVAE-TTS are designed to capture phoneme-duration dependencies. This focus provides the decoder with a nuanced understanding of how the phoneme contributes to the overall speech utterance, which can be interpreted by the inference fundamental frequency (F0) in \cref{fig.Exp2_f0}. We provide detailed additional interpretations in \cref{sec:F0_compare}.

% \textit{Detail interpretations in supplementary.}
% Even though BDPVAE-TTS gets slightly higher MCD than~\cite{lu2021vaenar}, it obtains discrete monotonic alignment which provides the decoder a better understanding of how each phoneme contributes to the overall acoustic character. This can be interpreted by the inference fundamental frequency (F0) in \cref{fig.Exp2_f0}.
% The inference fundamental frequency (F0) in  \cref{fig.Exp2_f0} shows that the intonation of our method is closer to the ground truth.
% We provide detailed additional interpretations of this experiment in the supplementary.

\subsection{Application: End-to-end Singing Voice Synthesis}

We extend the BDP-VAE with MA for end-to-end SVS on the popcs dataset~\citep{liu2022diffsinger}.
We perform this experiment not to compare with other models as in \cref{sec:tts}, but to demonstrate the utility of our method in a related task. In SVS, the longer phoneme duration also provides an opportunity to visualize the monotonically aligned optimal path clearly.
The popcs contains 117 Chinese Mandarin pop songs (5 hours) collected from a qualified female vocalist. We randomly split 50 clips for inference and the rest for training. During the inference phase, the conditional inputs are the fundamental frequency and lyrics of the song clips.

Similar to TTS task, SVS task also involves unobserved structural dependencies. BDP-VAE framework could help the SVS task to capture the discrete structural dependencies in parallel, leading to an end to end framework and a better model interpretability.

Two inference results are visualized in \cref{fig.Exp3_spect} where red rectangles indicate that the temporal structure between the generated Mel-spectrogram and ground truth is almost identical.
As expected, the decoder of BDP-VAE synthesizes spectrograms according to the extended conditions (i.e., the phonemes) mapping by the dependency of the latent monotonic path from the prior encoder.
Therefore, the decoder synthesizes singing voice according to latent path-aligned phoneme conditions.

% the latent monotonic path of the prior encoder maps the information of conditions (i.e., the phonemes) to generate spectrograms according to the 
% governs the dependency between conditions (i.e., the phonemes) and spectrograms to be generated by the decoder during inference. 
% Thus the decoder synthesizes singing voice according to latent path-aligned phoneme conditions.
% the latent path aligns conditions by 
% This figure shows that, as expected, the latent monotonically aligned path from the prior encoder governs the structure of the phoneme spectrograms generated by the decoder. The latent monotonic optimal path helps the decoder understand the structure of the unknown spectrogram from phoneme tokens.

\begin{table} [t]
  \caption{Mean absolute error and standard deviation (MAE/MAE$\pm$STD) of phoneme duration on TIMIT.}
  \footnotesize
  \renewcommand{\tabcolsep}{1mm}
  \label{MAD-TIMIT}
  % \vskip 0.05in
  \centering
  \begin{tabular}{llll}
  % \footnotesize
  % \renewcommand{\arraystretch}{0.9}
  % \renewcommand{\tabcolsep}{0.8mm}
    \toprule
    \textbf{Model} & \textbf{Distribution} & \textbf{Train} & \textbf{Inference} \\
    \midrule
    BDP-VAE  & $\mathcal{D}(\mathcal{R}_{\text{DTW}},\text{NN}_{\{\phi/\theta\}}, 1)$ & 2.92 &  3.93 $\pm$ 0.37 \\
    Baseline &  $\mathcal{D}(\mathcal{R}_{\text{DTW}},\mathbf{W}_{U(0,1)}, 1)$ & 5.67 &  5.69 $\pm$ 0.31 \\
    \bottomrule
  \end{tabular}
  % \vskip -0.2in
\end{table}

\begin{figure}[t]
% \vskip 0.2in
\begin{center}
\centerline{\includegraphics[width=0.99\columnwidth]{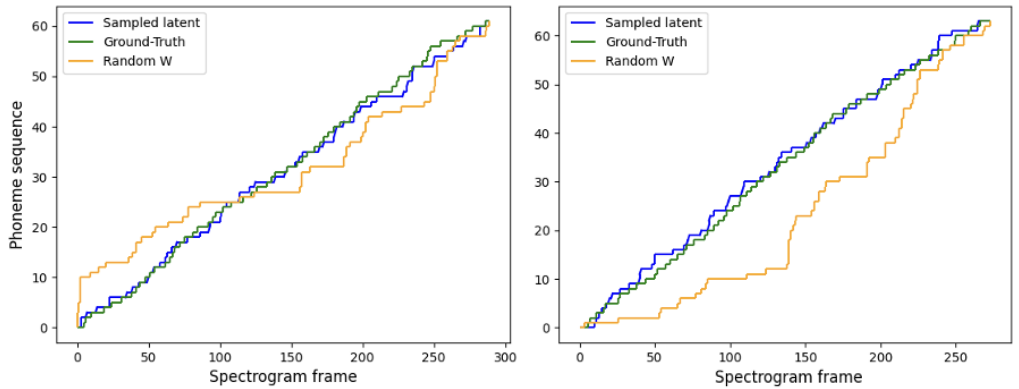}}
\caption{Visualization of GT alignment between phoneme tokens and spectrogram frames, latent optimal paths from the encoder, optimal paths from random latent space for two audio clips. \textit{BDP-VAE achieves closer alignments with GT, indicating its effectiveness in finding latent optimal paths.}}
\label{fig.vis_align}
\end{center}
\vskip -0.2in
\end{figure}

\subsection{Verify Behaviour of Latent Optimal Path in BDP-VAE} \label{sec:lop}

To verify the behaviour and generalization of the latent optimal paths in BDP-VAE, we obtain latent optimal paths under the computational graph of DTW on the TIMIT speech corpus dataset~\citep{SWVENO_1993} which includes manually time-aligned phonetic and word transcriptions.
TIMIT contains English speech of 630 speakers which
% in eight major dialects of American English.
utterances are recorded as 16-bit 16kHz speech waveform files. We randomly split 50 clips for testing and 580 clips for training. 

To the best of our knowledge, there are no existing similar works that enable VAEs to obtain hard latent optimal paths given any defined DAG structures. Thus, inspired by the experimental designs in ~\citet{mensch2018differentiable} and ~\citet{van2017neural}, we set a latent distribution $\mathcal{D}(\mathcal{R}_{\text{DTW}},\mathbf{W} = \{ w \in \mathbf{W} \sim U(0,1)\},1)$ as the baseline due to lack of direct comparison. The random baseline still follows the DTW constraint with uniform edge weights, which can verify the behaviour of latent space in BPD-VAE.  

We take a summation along the spectrogram dimension to obtain phoneme duration and compute the mean absolute error (MAE) between the ground truth and the phoneme duration. We input phoneme tokens and spectrogram lengths as conditions on inference to obtain latent optimal paths of the prior encoder. We repeat the inference 5 times and take the average and standard deviation of MAEs as our metric of evaluation. 
As in \cref{MAD-TIMIT}, our MAE is lower than the baseline 
% the MAEs of latent optimal paths are lower than the baseline 
indicating that the BDP-VAE captures meaningful information to obtain stochastic optimal paths in the latent space and is not merely guessing randomly. \cref{fig.vis_align} visualizes ground-truth alignments, latent optimal paths from BDP-VAE, and optimal paths from the baseline latent distribution on two audio clips, which clearly shows the latent optimal path from BDP-VAE under DTW gets a close alignment to the GT. This indicates BDP-VAE has the ability to learn informative edge weights $\mathbf{W}$ and 
capture sparse optimal paths in latent space. 

\subsection{Sensitivity of Hyper-parameter in BDP-VAE}\label{F:discussion}

\begin{table} [t]
  \caption{Mean absolute error and standard deviation (MAE / MAE $\pm$ STD) for the duration on TIMIT with different hyper-parameter $\alpha$ settings.}
  \footnotesize
  \vskip -0.05in
  \label{MAD-TIMIT-alpha}
  \centering
  \begin{tabular}{lllll}
    \toprule
    \textbf{Model} &   \textbf{Alpha}  & \textbf{Train} & \textbf{Inference} \\
    \midrule
    BDP-VAE  & 0.5  & 2.99 & 4.40 $\pm$ 0.46  \\
    BDP-VAE  & 1 & 2.92 & 3.93 $\pm$ 0.37\\
    % BDP-VAE  &  2.94 &  3.89 $\pm$ 0.35 \\
    BDP-VAE  & 1.5 & 2.94 & 3.89 $\pm$ 0.35 \\
    BDP-VAE & 2 & 3.01  & 4.08 $\pm$ 0.27\\
    BDP-VAE  & 3& 3.25  &  4.53 $\pm$ 0.10\\
    % Random W    &  5.67 &  5.69 $\pm$ 0.31 \\
    \bottomrule
  \end{tabular}
  \vskip -0.2in
\end{table}

To study the sensitivity of the temperature parameter $\alpha$, we extend the experiment in \cref{sec:lop}. \cref{MAD-TIMIT-alpha} shows the MAE value on the training and inference phase with different $\alpha$ settings. 
As discussed in \cref{sec:toy}, the $\alpha$ affects the sharpness of the path distribution. When $\alpha$ is smaller, the distribution is more stochastic. As $\alpha$ increases, the distribution becomes sharper. However, when the $\alpha$ value becomes large, the latent path alignment performance decreases, which is consistent with the contribution of temperature in~\citet{PlattProbabilisticOutputs1999}. 
% As $\alpha$ increases, the distribution becomes sharper, while the distribution of paths becomes uniform if the alpha is close to 0. 
In BDP-VAE, the model with too large $\alpha$ may not capture enough variety of data, conversely, for too small $\alpha$ the model becomes too spread out and lacks meaningful structure, with the distribution approaching uniform as $\alpha \rightarrow 0$. In real applications, the value of $\alpha$ in BDP-VAE should be located in a reasonable range which could be found by tuning or setting a learnable parameter.

\section{Conclusion} \label{sec:conclusion}

% We present a method that captures sparse latent optimal paths relying on the VAE framework. To this end, w
We introduce a probabilistic softening solution to the classical optimal path problem on DAGs, as stochastic optimal paths.
To achieve variational Bayesian inference with latent paths, we give efficient and tractable algorithms for sampling, likelihood and KL divergence within the family of path distributions in linear time with respect to edge numbers by dynamic programming with properties of the Gumbel distribution as message-passing, namely Bayesian dynamic programming with Gumbel propagation.
% and give algorithms for variational Bayesian inference. 
We demonstrate the usage of stochastic optimal paths in VAE framework and propose BDP-VAE. BDP-VAE captures sparse optimal paths as latent representations given a DAG and further achieves end-to-end training for downstream generative tasks that rely on unobserved structural relationships.
We showed how the BDP finds stochastic optimal paths under general small toy DAGs and demonstrated the BDP-VAE with the computational graph of monotonic alignment on two real-world applications to achieve an end-to-end framework.
We verified the behaviour and generalization of the latent optimal paths under the computational graph of dynamic time warping. We also studied the sensitivity of the hyper-parameter $\alpha$ and gave suggestions for real applications. Our experiments show the success of our approach on generative tasks where it achieves end-to-end training involving unobserved sparse structural optimal paths.
Beyond VAE, the BDP can potentially be integrated into other probabilistic frameworks or planning in model-based reinforcement learning to obtain latent paths that leave for future extensions.\\
% \noindent\textbf{Limitations and Broader Impacts:}
\noindent\textbf{Limitations:}
As a discrete latent variable model, BDP-VAE uses the REINFORCE estimator which may lead to high gradient variance and slow convergence during training. This limitation may be solved by involving variance reduction techniques during training.

\section*{Impact Statement}

This paper presents work whose goal is to advance the field of Machine Learning and applications of its downstream tasks. There are many potential societal consequences of our work; however, we do not foresee our methods bringing negative social impacts.


% \bibliography{example_paper}

\begin{thebibliography}{46}
\providecommand{\natexlab}[1]{#1}
\providecommand{\url}[1]{\texttt{#1}}
\expandafter\ifx\csname urlstyle\endcsname\relax
  \providecommand{\doi}[1]{doi: #1}\else
  \providecommand{\doi}{doi: \begingroup \urlstyle{rm}\Url}\fi

\bibitem[Amos \& Kolter(2017)Amos and Kolter]{AmosK17}
Amos, B. and Kolter, J.~Z.
\newblock Optnet: Differentiable optimization as a layer in neural networks.
\newblock In Precup, D. and Teh, Y.~W. (eds.), \emph{Proceedings of the 34th International Conference on Machine Learning, {ICML} 2017, Sydney, NSW, Australia, 6-11 August 2017}, volume~70 of \emph{Proceedings of Machine Learning Research}, pp.\  136--145. {PMLR}, 2017.
\newblock URL \url{http://proceedings.mlr.press/v70/amos17a.html}.

\bibitem[BakIr et~al.(2007)BakIr, Hofmann, Smola, Sch{\"o}lkopf, and Taskar]{bakir2007predicting}
BakIr, G., Hofmann, T., Smola, A.~J., Sch{\"o}lkopf, B., and Taskar, B.
\newblock \emph{Predicting structured data}.
\newblock MIT press, 2007.

\bibitem[Cai et~al.(2019)Cai, Xu, Yi, Huang, and Rajasekaran]{cai2019dtwnet}
Cai, X., Xu, T., Yi, J., Huang, J., and Rajasekaran, S.
\newblock Dtwnet: a dynamic time warping network.
\newblock \emph{Advances in neural information processing systems}, 32, 2019.

\bibitem[Chen et~al.(2022)Chen, Tan, Qi, Zhou, Li, and Wu]{ChenTQZLW22}
Chen, Q., Tan, M., Qi, Y., Zhou, J., Li, Y., and Wu, Q.
\newblock {V2C:} visual voice cloning.
\newblock In \emph{{IEEE/CVF} Conference on Computer Vision and Pattern Recognition, {CVPR} 2022, New Orleans, LA, USA, June 18-24, 2022}, pp.\  21210--21219. {IEEE}, 2022.
\newblock \doi{10.1109/CVPR52688.2022.02056}.
\newblock URL \url{https://doi.org/10.1109/CVPR52688.2022.02056}.

\bibitem[Chiu \& Raffel(2018)Chiu and Raffel]{ChiuR18}
Chiu, C. and Raffel, C.
\newblock Monotonic chunkwise attention.
\newblock In \emph{6th International Conference on Learning Representations, {ICLR} 2018, Vancouver, BC, Canada, April 30 - May 3, 2018, Conference Track Proceedings}. OpenReview.net, 2018.
\newblock URL \url{https://openreview.net/forum?id=Hko85plCW}.

\bibitem[Deng et~al.(2018)Deng, Kim, Chiu, Guo, and Rush]{DengKCGR18}
Deng, Y., Kim, Y., Chiu, J.~T., Guo, D., and Rush, A.~M.
\newblock Latent alignment and variational attention.
\newblock In Bengio, S., Wallach, H.~M., Larochelle, H., Grauman, K., Cesa{-}Bianchi, N., and Garnett, R. (eds.), \emph{Advances in Neural Information Processing Systems 31: Annual Conference on Neural Information Processing Systems 2018, NeurIPS 2018, December 3-8, 2018, Montr{\'{e}}al, Canada}, pp.\  9735--9747, 2018.
\newblock URL \url{https://proceedings.neurips.cc/paper/2018/hash/b691334ccf10d4ab144d672f7783c8a3-Abstract.html}.

\bibitem[Djolonga \& Krause(2017)Djolonga and Krause]{Djolonga017}
Djolonga, J. and Krause, A.
\newblock Differentiable learning of submodular functions.
\newblock In Guyon, I., von Luxburg, U., Bengio, S., Wallach, H.~M., Fergus, R., Vishwanathan, S. V.~N., and Garnett, R. (eds.), \emph{Advances in Neural Information Processing Systems 30: Annual Conference on Neural Information Processing Systems 2017, December 4-9, 2017, Long Beach, CA, {USA}}, pp.\  1013--1023, 2017.
\newblock URL \url{https://proceedings.neurips.cc/paper/2017/hash/192fc044e74dffea144f9ac5dc9f3395-Abstract.html}.

\bibitem[Garofolo et~al.(1992)Garofolo, Lamel, Fisher, Fiscus, Pallett, Dahlgren, and Zue]{SWVENO_1993}
Garofolo, J., Lamel, L., Fisher, W., Fiscus, J., Pallett, D., Dahlgren, N., and Zue, V.
\newblock Timit acoustic-phonetic continuous speech corpus.
\newblock \emph{Linguistic Data Consortium}, 11 1992.

\bibitem[Halperin et~al.(2019)Halperin, Ephrat, and Peleg]{halperin2019dynamic}
Halperin, T., Ephrat, A., and Peleg, S.
\newblock Dynamic temporal alignment of speech to lips.
\newblock In \emph{ICASSP 2019-2019 IEEE International Conference on Acoustics, Speech and Signal Processing (ICASSP)}, pp.\  3980--3984. IEEE, 2019.

\bibitem[Hasegawa-Johnson et~al.(2005)Hasegawa-Johnson, Cole, Hirschberg, Jilka, and Tannenbaum]{johnson2005penn}
Hasegawa-Johnson, M., Cole, J., Hirschberg, J., Jilka, M., and Tannenbaum, R.
\newblock Penn phonetics toolkit (p2tk): A software suite for sound analysis as a function of time.
\newblock Technical report, Department of Linguistics, University of Pennsylvania, 2005.

\bibitem[Heckerman(1998)]{Heckerman98}
Heckerman, D.
\newblock A tutorial on learning with bayesian networks.
\newblock In Jordan, M.~I. (ed.), \emph{Learning in Graphical Models}, volume~89 of \emph{{NATO} {ASI} Series}, pp.\  301--354. Springer Netherlands, 1998.
\newblock \doi{10.1007/978-94-011-5014-9\_11}.
\newblock URL \url{https://doi.org/10.1007/978-94-011-5014-9\_11}.

\bibitem[Jang et~al.(2017)Jang, Gu, and Poole]{JangGP17}
Jang, E., Gu, S., and Poole, B.
\newblock Categorical reparameterization with gumbel-softmax.
\newblock In \emph{5th International Conference on Learning Representations, {ICLR} 2017, Toulon, France, April 24-26, 2017, Conference Track Proceedings}. OpenReview.net, 2017.
\newblock URL \url{https://openreview.net/forum?id=rkE3y85ee}.

\bibitem[Jeong et~al.(2021)Jeong, Kim, Cheon, Choi, and Kim]{JeongKCCK21}
Jeong, M., Kim, H., Cheon, S.~J., Choi, B.~J., and Kim, N.~S.
\newblock Diff-tts: {A} denoising diffusion model for text-to-speech.
\newblock In Hermansky, H., Cernock{\'{y}}, H., Burget, L., Lamel, L., Scharenborg, O., and Motl{\'{\i}}cek, P. (eds.), \emph{Interspeech 2021, 22nd Annual Conference of the International Speech Communication Association, Brno, Czechia, 30 August - 3 September 2021}, pp.\  3605--3609. {ISCA}, 2021.
\newblock \doi{10.21437/Interspeech.2021-469}.
\newblock URL \url{https://doi.org/10.21437/Interspeech.2021-469}.

\bibitem[Kim et~al.(2020)Kim, Kim, Kong, and Yoon]{kim2020glow}
Kim, J., Kim, S., Kong, J., and Yoon, S.
\newblock Glow-tts: A generative flow for text-to-speech via monotonic alignment search.
\newblock \emph{Advances in Neural Information Processing Systems}, 33:\penalty0 8067--8077, 2020.

\bibitem[Kingma \& Dhariwal(2018)Kingma and Dhariwal]{kingma2018glow}
Kingma, D.~P. and Dhariwal, P.
\newblock Glow: Generative flow with invertible 1x1 convolutions.
\newblock \emph{Advances in neural information processing systems}, 31, 2018.

\bibitem[Kingma \& Welling(2014)Kingma and Welling]{KingmaW13}
Kingma, D.~P. and Welling, M.
\newblock Auto-encoding variational bayes.
\newblock In Bengio, Y. and LeCun, Y. (eds.), \emph{2nd International Conference on Learning Representations, {ICLR} 2014, Banff, AB, Canada, April 14-16, 2014, Conference Track Proceedings}, 2014.
\newblock URL \url{http://arxiv.org/abs/1312.6114}.

\bibitem[Kubichek(1993)]{kubichek1993mel}
Kubichek, R.
\newblock Mel-cepstral distance measure for objective speech quality assessment.
\newblock In \emph{Proceedings of IEEE pacific rim conference on communications computers and signal processing}, volume~1, pp.\  125--128. IEEE, 1993.

\bibitem[Li et~al.(2022)Li, Zhao, Zhelun, and Sheng]{li2022danceformer}
Li, B., Zhao, Y., Zhelun, S., and Sheng, L.
\newblock Danceformer: Music conditioned 3d dance generation with parametric motion transformer.
\newblock In \emph{Proceedings of the AAAI Conference on Artificial Intelligence}, pp.\  1272--1279, 2022.

\bibitem[Li et~al.(2018)Li, Liu, Liu, Zhao, Liu, and Zhou]{abs-1809-08895}
Li, N., Liu, S., Liu, Y., Zhao, S., Liu, M., and Zhou, M.
\newblock Close to human quality {TTS} with transformer.
\newblock \emph{CoRR}, abs/1809.08895, 2018.
\newblock URL \url{http://arxiv.org/abs/1809.08895}.

\bibitem[Liu et~al.(2022)Liu, Li, Ren, Chen, and Zhao]{liu2022diffsinger}
Liu, J., Li, C., Ren, Y., Chen, F., and Zhao, Z.
\newblock Diffsinger: Singing voice synthesis via shallow diffusion mechanism.
\newblock In \emph{Proceedings of the AAAI Conference on Artificial Intelligence}, pp.\  11020--11028, 2022.

\bibitem[Lu et~al.(2021)Lu, Wu, Wu, Li, Kang, Liu, and Meng]{lu2021vaenar}
Lu, H., Wu, Z., Wu, X., Li, X., Kang, S., Liu, X., and Meng, H.
\newblock Vaenar-tts: Variational auto-encoder based non-autoregressive text-to-speech synthesis.
\newblock \emph{arXiv preprint arXiv:2107.03298}, 2021.

\bibitem[Ma et~al.(2019)Ma, Zhou, Li, Neubig, and Hovy]{flowseq}
Ma, X., Zhou, C., Li, X., Neubig, G., and Hovy, E.
\newblock {F}low{S}eq: Non-autoregressive conditional sequence generation with generative flow.
\newblock In \emph{Proceedings of the 2019 Conference on Empirical Methods in Natural Language Processing and the 9th International Joint Conference on Natural Language Processing (EMNLP-IJCNLP)}, pp.\  4282--4292, Hong Kong, China, November 2019. Association for Computational Linguistics.
\newblock \doi{10.18653/v1/D19-1437}.
\newblock URL \url{https://aclanthology.org/D19-1437}.

\bibitem[Maddison et~al.(2014)Maddison, Tarlow, and Minka]{maddison2014sampling}
Maddison, C.~J., Tarlow, D., and Minka, T.
\newblock A* sampling.
\newblock \emph{Advances in neural information processing systems}, 27, 2014.

\bibitem[Maddison et~al.(2017)Maddison, Mnih, and Teh]{MaddisonMT17}
Maddison, C.~J., Mnih, A., and Teh, Y.~W.
\newblock The concrete distribution: {A} continuous relaxation of discrete random variables.
\newblock In \emph{5th International Conference on Learning Representations, {ICLR} 2017, Toulon, France, April 24-26, 2017, Conference Track Proceedings}. OpenReview.net, 2017.
\newblock URL \url{https://openreview.net/forum?id=S1jE5L5gl}.

\bibitem[McAuliffe et~al.(2017)McAuliffe, Socolof, Mihuc, Wagner, and Sonderegger]{McAuliffeSM0S17}
McAuliffe, M., Socolof, M., Mihuc, S., Wagner, M., and Sonderegger, M.
\newblock Montreal forced aligner: Trainable text-speech alignment using kaldi.
\newblock In Lacerda, F. (ed.), \emph{Interspeech 2017, 18th Annual Conference of the International Speech Communication Association, Stockholm, Sweden, August 20-24, 2017}, pp.\  498--502. {ISCA}, 2017.
\newblock URL \url{http://www.isca-speech.org/archive/Interspeech\_2017/abstracts/1386.html}.

\bibitem[Mehrish et~al.(2023)Mehrish, Majumder, Bharadwaj, Mihalcea, and Poria]{mehrish2023review}
Mehrish, A., Majumder, N., Bharadwaj, R., Mihalcea, R., and Poria, S.
\newblock A review of deep learning techniques for speech processing.
\newblock \emph{Information Fusion}, pp.\  101869, 2023.

\bibitem[Mensch \& Blondel(2018)Mensch and Blondel]{mensch2018differentiable}
Mensch, A. and Blondel, M.
\newblock Differentiable dynamic programming for structured prediction and attention.
\newblock In \emph{International Conference on Machine Learning}, pp.\  3462--3471. PMLR, 2018.

\bibitem[Mohamed et~al.(2020)Mohamed, Rosca, Figurnov, and Mnih]{mohamed2020monte}
Mohamed, S., Rosca, M., Figurnov, M., and Mnih, A.
\newblock Monte carlo gradient estimation in machine learning.
\newblock \emph{J. Mach. Learn. Res.}, 21\penalty0 (132):\penalty0 1--62, 2020.

\bibitem[Peng et~al.(2020)Peng, Ping, Song, and Zhao]{peng2020non}
Peng, K., Ping, W., Song, Z., and Zhao, K.
\newblock Non-autoregressive neural text-to-speech.
\newblock In \emph{International conference on machine learning}, pp.\  7586--7598. PMLR, 2020.

\bibitem[Petrov \& Klein(2007)Petrov and Klein]{PetrovK07}
Petrov, S. and Klein, D.
\newblock Discriminative log-linear grammars with latent variables.
\newblock In Platt, J.~C., Koller, D., Singer, Y., and Roweis, S.~T. (eds.), \emph{Advances in Neural Information Processing Systems 20, Proceedings of the Twenty-First Annual Conference on Neural Information Processing Systems, Vancouver, British Columbia, Canada, December 3-6, 2007}, pp.\  1153--1160. Curran Associates, Inc., 2007.
\newblock URL \url{https://proceedings.neurips.cc/paper/2007/hash/9cc138f8dc04cbf16240daa92d8d50e2-Abstract.html}.

\bibitem[Platt(2000)]{PlattProbabilisticOutputs1999}
Platt, J.
\newblock Probabilistic outputs for support vector machines and comparison to regularized likelihood methods.
\newblock In \emph{Advances in Large Margin Classifiers}, 2000.

\bibitem[Popov et~al.(2021)Popov, Vovk, Gogoryan, Sadekova, and Kudinov]{popov2021grad}
Popov, V., Vovk, I., Gogoryan, V., Sadekova, T., and Kudinov, M.
\newblock Grad-tts: A diffusion probabilistic model for text-to-speech.
\newblock In \emph{International Conference on Machine Learning}, pp.\  8599--8608. PMLR, 2021.

\bibitem[Rabiner(1989)]{Rabiner89}
Rabiner, L.~R.
\newblock A tutorial on hidden markov models and selected applications in speech recognition.
\newblock \emph{Proc. {IEEE}}, 77\penalty0 (2):\penalty0 257--286, 1989.
\newblock \doi{10.1109/5.18626}.
\newblock URL \url{https://doi.org/10.1109/5.18626}.

\bibitem[Ren et~al.(2019)Ren, Ruan, Tan, Qin, Zhao, Zhao, and Liu]{ren2019fastspeech}
Ren, Y., Ruan, Y., Tan, X., Qin, T., Zhao, S., Zhao, Z., and Liu, T.-Y.
\newblock Fastspeech: Fast, robust and controllable text to speech.
\newblock \emph{Advances in Neural Information Processing Systems}, 32, 2019.

\bibitem[Ren et~al.(2020)Ren, Hu, Tan, Qin, Zhao, Zhao, and Liu]{ren2020fastspeech}
Ren, Y., Hu, C., Tan, X., Qin, T., Zhao, S., Zhao, Z., and Liu, T.-Y.
\newblock Fastspeech 2: Fast and high-quality end-to-end text to speech.
\newblock \emph{arXiv preprint arXiv:2006.04558}, 2020.

\bibitem[Sakoe \& Chiba(1978)Sakoe and Chiba]{sakoe1978dynamic}
Sakoe, H. and Chiba, S.
\newblock Dynamic programming algorithm optimization for spoken word recognition.
\newblock \emph{IEEE transactions on acoustics, speech, and signal processing}, 26\penalty0 (1):\penalty0 43--49, 1978.

\bibitem[Shen et~al.(2018{\natexlab{a}})Shen, Pang, Weiss, Schuster, Jaitly, Yang, Chen, Zhang, Wang, Ryan, Saurous, Agiomyrgiannakis, and Wu]{tacotron2}
Shen, J., Pang, R., Weiss, R.~J., Schuster, M., Jaitly, N., Yang, Z., Chen, Z., Zhang, Y., Wang, Y., Ryan, R., Saurous, R.~A., Agiomyrgiannakis, Y., and Wu, Y.
\newblock Natural {TTS} synthesis by conditioning wavenet on {MEL} spectrogram predictions.
\newblock In \emph{2018 {IEEE} International Conference on Acoustics, Speech and Signal Processing, {ICASSP} 2018, Calgary, AB, Canada, April 15-20, 2018}, pp.\  4779--4783. {IEEE}, 2018{\natexlab{a}}.
\newblock \doi{10.1109/ICASSP.2018.8461368}.
\newblock URL \url{https://doi.org/10.1109/ICASSP.2018.8461368}.

\bibitem[Shen et~al.(2018{\natexlab{b}})Shen, Pang, Weiss, Schuster, Jaitly, Yang, Chen, Zhang, Wang, Skerrv-Ryan, et~al.]{postnet}
Shen, J., Pang, R., Weiss, R.~J., Schuster, M., Jaitly, N., Yang, Z., Chen, Z., Zhang, Y., Wang, Y., Skerrv-Ryan, R., et~al.
\newblock Natural tts synthesis by conditioning wavenet on mel spectrogram predictions.
\newblock In \emph{2018 IEEE international conference on acoustics, speech and signal processing (ICASSP)}, pp.\  4779--4783. IEEE, 2018{\natexlab{b}}.

\bibitem[Struminsky et~al.(2021)Struminsky, Gadetsky, Rakitin, Karpushkin, and Vetrov]{StruminskyGRKV21}
Struminsky, K., Gadetsky, A., Rakitin, D., Karpushkin, D., and Vetrov, D.~P.
\newblock Leveraging recursive gumbel-max trick for approximate inference in combinatorial spaces.
\newblock In Ranzato, M., Beygelzimer, A., Dauphin, Y.~N., Liang, P., and Vaughan, J.~W. (eds.), \emph{Advances in Neural Information Processing Systems 34: Annual Conference on Neural Information Processing Systems 2021, NeurIPS 2021, December 6-14, 2021, virtual}, pp.\  10999--11011, 2021.
\newblock URL \url{https://proceedings.neurips.cc/paper/2021/hash/5b658d2a925565f0755e035597f8d22f-Abstract.html}.

\bibitem[Tralie \& Dempsey(2020)Tralie and Dempsey]{tralie2020exact}
Tralie, C. and Dempsey, E.
\newblock Exact, parallelizable dynamic time warping alignment with linear memory.
\newblock \emph{arXiv preprint arXiv:2008.02734}, 2020.

\bibitem[Van Den~Oord et~al.(2017)Van Den~Oord, Vinyals, et~al.]{van2017neural}
Van Den~Oord, A., Vinyals, O., et~al.
\newblock Neural discrete representation learning.
\newblock \emph{Advances in neural information processing systems}, 30, 2017.

\bibitem[Vaswani et~al.(2017)Vaswani, Shazeer, Parmar, Uszkoreit, Jones, Gomez, Kaiser, and Polosukhin]{vaswani2017attention}
Vaswani, A., Shazeer, N., Parmar, N., Uszkoreit, J., Jones, L., Gomez, A.~N., Kaiser, {\L}., and Polosukhin, I.
\newblock Attention is all you need.
\newblock \emph{Advances in neural information processing systems}, 30, 2017.

\bibitem[Verdu \& Poor(1987)Verdu and Poor]{verdu1987abstract}
Verdu, S. and Poor, H.~V.
\newblock Abstract dynamic programming models under commutativity conditions.
\newblock \emph{SIAM Journal on Control and Optimization}, 25\penalty0 (4):\penalty0 990--1006, 1987.

\bibitem[Yu et~al.(2016{\natexlab{a}})Yu, Blunsom, Dyer, Grefenstette, and Kocisky]{yu2016neural}
Yu, L., Blunsom, P., Dyer, C., Grefenstette, E., and Kocisky, T.
\newblock The neural noisy channel.
\newblock \emph{arXiv preprint arXiv:1611.02554}, 2016{\natexlab{a}}.

\bibitem[Yu et~al.(2016{\natexlab{b}})Yu, Buys, and Blunsom]{yu2016a}
Yu, L., Buys, J., and Blunsom, P.
\newblock Online segment to segment neural transduction.
\newblock \emph{arXiv preprint arXiv:1609.08194}, 2016{\natexlab{b}}.

\bibitem[Zandie et~al.(2021)Zandie, Mahoor, Madsen, and Emamian]{zandie2021ryanspeech}
Zandie, R., Mahoor, M.~H., Madsen, J., and Emamian, E.~S.
\newblock Ryanspeech: A corpus for conversational text-to-speech synthesis.
\newblock \emph{ISCA}, 2021.
\newblock \doi{10.21437/Interspeech.2021-341}.
\newblock URL \url{https://doi.org/10.21437/Interspeech.2021-341}.

\end{thebibliography}
% \bibliographystyle{icml2024}
{
\bibliographystyle{unsrt}

}

%%%%%%%%%%%%%%%%%%%%%%%%%%%%%%%%%%%%%%%%%%%%%%%%%%%%%%%%%%%%%%%%%%%%%%%%%%%%%%%
%%%%%%%%%%%%%%%%%%%%%%%%%%%%%%%%%%%%%%%%%%%%%%%%%%%%%%%%%%%%%%%%%%%%%%%%%%%%%%%
% APPENDIX
%%%%%%%%%%%%%%%%%%%%%%%%%%%%%%%%%%%%%%%%%%%%%%%%%%%%%%%%%%%%%%%%%%%%%%%%%%%%%%%
%%%%%%%%%%%%%%%%%%%%%%%%%%%%%%%%%%%%%%%%%%%%%%%%%%%%%%%%%%%%%%%%%%%%%%%%%%%%%%%
\newpage
\appendix
\onecolumn
\section{Proofs of each lemma in Gumbel propagation} \label{A:gumbel_proof}
\subsection{Proofs for \textbf{Lemma 4.2}}

\proof{
The result follows directly from \eqref{eqn:gumbelmax} and \eqref{eqn:gumbelshift}.
\qed
}

\subsection{Proofs for \textbf{Lemma 4.3}}
\proof{
The result follows directly from \eqref{eqn:gumbelmax} and \eqref{eqn:gumbelshift}. \qed
}

\subsection{Proofs for \textbf{Lemma 4.4}}

\proof{
From the DAG structure of $\mathcal R$ we have
\begin{align}
\label{eqn:dagrecursion}
	\mathcal Y(1,v) = \bigcup_{u\in\mathcal P(v)} \left\{\mathbf y\cdot v:  \forall \mathbf y \in \mathcal Y(1,u) \right\},
\end{align}
where $\mathbf y\cdot v =(y_1,y_2,\dots,y_{\abs{y}}, v)$ denotes concatenation. 

Then we have
\begin{align}
\label{eqn:aa}
	\mu_v 
	& = \log \sum_{\mathbf y\in\mathcal Y(1,v)} \exp(\alpha\left\|\mathbf y\right\|_{\mathbf W})
	\\
\label{eqn:bb}
	& = \log \sum_{\mathbf y\in\bigcup_{u\in\mathcal P(v)} \left\{\widehat{\mathbf y}\cdot v:  \forall \widehat{\mathbf y} \in \mathcal Y(1,u) \right\}} \exp(\alpha\left\|\mathbf y\right\|_{\mathbf W})
	\\
\label{eqn:cc}
	& = \log \sum_{u\in\mathcal P(v)} \sum_{\widehat{\mathbf y}\in \mathcal Y(1,u)} \exp(\alpha\left\|\widehat{\mathbf y}\cdot v\right\|_{\mathbf W}) \\
\label{eqn:dd}
	& = \log \sum_{u\in\mathcal P(v)} \sum_{\widehat{\mathbf y}\in \mathcal Y(1,u)} \exp\Big(\alpha\left\|\widehat{\mathbf y}\right\|_{\mathbf W}+\alpha\, w_{u,v}\Big) \\
\label{eqn:ee}
	& = \log \sum_{u\in\mathcal P(v)} \exp\Big(\log\sum_{\widehat{\mathbf y}\in \mathcal Y(1,u)}\exp\big(\alpha\left\|\widehat{\mathbf y}\right\|_{\mathbf W}\big)+\alpha\, w_{u,v}\Big) \\
\label{eqn:ff}
	& = \log \sum_{u\in\mathcal P(v)} \exp\Big(\mu_u+\alpha\, w_{u,v}\Big),
\end{align}

where \cref{eqn:aa} restates \eqref{eqn:mudefinition}, \cref{eqn:bb} follows from \eqref{eqn:dagrecursion}, 
\cref{eqn:cc} expands the summation, \cref{eqn:dd} follows from the definition of $\left\|\mathbf y\right\|_{\mathbf W}$,
% where 
\cref{eqn:ee} follows from the identity 

\begin{align}
\label{eqn:logsumexpidentity}
 	\sum_i\exp(a+b_i)=\exp(a+\log\sum_i \exp(b_i)),
 \end{align}
 
 \cref{eqn:ff} follows from 
 \begin{align}
     \mu_v = \log \sum_{\mathbf y\in\mathcal Y(1,v)} \exp(\alpha\left\|\mathbf y\right\|_{\mathbf W})
 \end{align}
 yielding \cref{eqn:murecursion} as required.
\qed
}

\section{Proofs of each lemma in Sampling and Likelihood} \label{B:sampling_likelihood_proof}

\subsection{Proofs for Lemma 4.5}
\proof{
	Assuming without loss of generality that $v=N$,
	\begin{align}
	\label{eqn:A}
		\pi_{u, v}
		& = 
		p(y_{i-1}=u|y_{i}=v,u\in\mathcal P(v))
		\\
\label{eqn:B}
		& = 
		\frac{p(y_{i-1}=u,y_{i}=v|u\in\mathcal P(v))}{p(y_{i}=v)}
		\\
		\label{eqn:C}
		& =\frac{1}{p(y_{i}=v)} \sum_{\widetilde{\mathbf y}\in \mathcal Y(1,u)} p(Y=\widetilde{\mathbf y}\cdot v)
		\\
		\label{eqn:D}
		& = \frac{1}{p(y_{i}=v)}
		\sum_{\widetilde{\mathbf y}\in \mathcal Y(1,u)} \frac{\exp(\alpha \left \| \widetilde {\mathbf y}\cdot v \right \|_{\mathbf W})}
	{\sum_{\widehat{\mathbf y}\in\mathcal Y(1,N)}\exp(\alpha \left \| \widehat {\mathbf y} \right \|_{\mathbf W})}
	\\
	\label{eqn:E}
			& \propto
		\sum_{\widetilde{\mathbf y} \in \mathcal Y(1,u)} \exp(\alpha \left \| \widetilde {\mathbf y}\right \|_{\mathbf W}+\alpha\, w_{u,v})
	\\
	\label{eqn:F}
	& \propto \exp\big(\log\sum_{\widetilde{\mathbf y} \in \mathcal Y(1,u)}\exp(\alpha \left \| \widetilde {\mathbf y}\right \|_{\mathbf W})+\alpha\, w_{u,v}\big)
	\\
	\label{eqn:G}
	& \propto \exp\big(\mu_u+\alpha\, w_{u,v}\big)
	\\
	\label{eqn:H}
	& = \frac{\exp\big(\mu_u+\alpha\, w_{u,v}\big)}{\sum_{\widetilde u\in\mathcal P(v)}\exp\big(\mu_{\widetilde u}+\alpha\, w_{\widetilde u,v}\big)}
	\\
	\label{eqn:I}
	& = \frac{\exp(\mu_u+\alpha\, w_{u,v})}{\exp(\mu_v)},
	\end{align}
% where \cref{eqn:C} marginalises over $\mathbf y_{<i-1}$, \cref{eqn:E} neglects factors that do not depend on $u$, \cref{eqn:H} makes the normalization explicit and \cref{eqn:I} uses \eqref{eqn:murecursion} to recover \eqref{eqn:transitionmu} as required. \qed

where \cref{eqn:B} rewrite \eqref{eqn:A} by the rule of conditional probability, \cref{eqn:C} marginalises over $\mathbf y_{<i-1}$, \cref{eqn:D} extend the probability by \cref{eqn:gibbsdistribution},
\cref{eqn:E} neglects factors that do not depend on $u$, \cref{eqn:F} uses the identity of \eqref{eqn:logsumexpidentity} then get \cref{eqn:G} according to 
\cref{eqn:mudefinition}.
\cref{eqn:H} makes the normalization explicit and \cref{eqn:I} uses \eqref{eqn:murecursion} to recover \cref{eqn:transitionmu} as required. \qed
}

\section{Proofs of each lemma in KL Divergence} \label{C:KL_proof}

\subsection{Proofs for Lemma 4.9}
\proof{
From the DAG structure of $\mathcal R$ we have, for all edges $(u,v)\in \mathcal E$,
\begin{align}
\label{eqn:dagdoublerecursion}
	\mathcal Y(1,v) = \bigcup_{\mathbf l\in\mathcal Y(1,u)} \bigcup_{\mathbf r\in\mathcal Y(v,N)} \mathbf l\cdot\mathbf r,
\end{align}
where we recall $\mathbf l\cdot\mathbf r$ denotes concatenation. We therefore have
\begin{align}
	\omega_{u,v} 
	& = 
	\sum_{\{\mathbf y \in\mathcal Y(1,N): (u,v)\in \mathbf y\}}\mathcal D(\mathbf y|\mathcal R, \mathbf W, \alpha)
	\label{eqn:omega:a}
	\\
	& = 
	\sum_{\{\mathbf y \in\mathcal Y(1,N): (u,v)\in \mathbf y\}}\prod_{(u',v')\in\mathbf y} \pi_{u',v'}
	\label{eqn:omega:b}
	\\
	& = 
	\sum_{\mathbf l\in\mathcal Y(1,u)} \sum_{\mathbf r\in\mathcal Y(v,N)} \prod_{(u',v')\in\mathbf l\cdot\mathbf r} \pi_{u',v'}
	\label{eqn:omega:c}
	\\
	& = 
	\sum_{\mathbf l\in\mathcal Y(1,u)} \sum_{\mathbf r\in\mathcal Y(v,N)} \pi_{u,v} \prod_{(u',v')\in\mathbf l} \pi_{u',v'} \prod_{(u',v')\in\mathbf r} \pi_{u'',v''}
	\label{eqn:omega:d}
	\\
	& = 
	\pi_{u,v} 
	\underbrace{\sum_{\mathbf l\in\mathcal Y(1,u)} \prod_{(u',v')\in\mathbf l} \pi_{u',v'}}_{\equiv \lambda_u}
	\underbrace{\sum_{\mathbf r\in\mathcal Y(v,N)} \prod_{(u',v')\in\mathbf r} \pi_{u'',v''}}_{\equiv \rho_v},
	\label{eqn:omega:e}
\end{align}

where \eqref{eqn:omega:a} restates \eqref{eqn:omegadefinition}, \eqref{eqn:omega:b} uses the chain rule of probability using \cref{eqn:pathchainrule},
\eqref{eqn:omega:c} follows from \eqref{eqn:dagdoublerecursion}, \eqref{eqn:omega:d} re-factors the product and the final \eqref{eqn:omega:e} rearranges sums and products.

Finally, it is straightforward to show by induction that the $\lambda_u$ and $\rho_v$ defined in \eqref{eqn:omega:e} above obey the classic sum-of-product recursions given in the statement of the lemma. For example,
\begin{align}
	\lambda_v
	& =
	\sum_{u\in\mathcal{P}(v)} \pi_{u,v} \lambda_u,
	\\
	& =
	\sum_{u\in\mathcal{P}(v)} \pi_{u,v} \sum_{\mathbf l\in\mathcal Y(1,u)} 
	\prod_{(u',v')\in\mathbf l} \pi_{u',v'}
	\\
	& =
	\sum_{u\in\mathcal{P}(v)} \sum_{\mathbf l\in\mathcal Y(1,u)} 
	\prod_{(u',v')\in (\mathbf l\cdot v)} \pi_{u',v'}
	\\
	& =
	\sum_{\mathbf l\in\mathcal Y(1,v)} \prod_{(u,v)\in\mathbf l} \pi_{u,v},
\end{align}
as required.
\qed
}

\subsection{Proofs for Lemma 4.10}
\proof{
We have

\begin{align} \label{eqn:kl_1}
    ~ & \kl{\mathcal{D}(\mathcal{R}, \mathbf{W}, \alpha)}{\mathcal{D}(\mathcal{R}, \mathbf{W}^{(r)}, \alpha)} \\
    & = 
    \theexpectation\left[
        \log \mathcal{D}(\mathbf{y}|\mathcal{R}, \mathbf{W}, \alpha)-\log \mathcal{D}(\mathbf{y}|\mathcal{R}, \mathbf{W}^{(r)}, \alpha)
    \right]
    \\
    & = 
    \theexpectation\Big[
        \log 
        \frac{\exp(\alpha \left \| \mathbf{y} \right \|_{\mathbf{W}})}
        {\sum_{\widehat {\mathbf{y}}\in\mathcal{Y}(1,N)}\exp(\alpha \left \| \widehat {\mathbf{y}} \right \|_{\mathbf{W}})}
    \nonumber
    \\
    & ~~~~~
        -\log
        \frac{\exp(\alpha \left \| \mathbf{y} \right \|_{\mathbf{W}^{(r)}})}
        {\sum_{\widehat {\mathbf{y}}\in\mathcal{Y}(1,N)}\exp(\alpha \left \| \widehat {\mathbf{y}} \right \|_{\mathbf{W}^{(r)}})}
    \Big] 
    \nonumber
    \\
    & = 
    \theexpectation\Big[
    \alpha \left \| \mathbf{y} \right \|_{\mathbf{W}}
    -
    \alpha \left \| \mathbf{y} \right \|_{\mathbf{W}^{(r)}}
    +
    \Big.
    \label{eqn:firsttwo}
    \\
    & ~~ \Big.
    \log
    \sum_{\widehat {\mathbf{y}}\in \mathcal{Y}(1,N)}\exp(\alpha \left \| \widehat {\mathbf{y}} \right \|_{\mathbf{W}^{(r)}})
    -
    \log
    \sum_{\widehat {\mathbf{y}}\in \mathcal{Y}(1,N)}\exp(\alpha \left \| \widehat {\mathbf{y}} \right \|_{\mathbf{W}})
    \Big].
    \label{eqn:secondtwo}
\end{align}
The third and fourth terms inside the final expectation (on line \eqref{eqn:secondtwo}) are easily handled; \textit{e.g.} for the third term we have

\begin{align} 
    \nonumber
    & \theexpectation\left[\log
    	\sum_{\widehat {\mathbf{y}}\in\mathcal{Y}(1,N)}\exp(\alpha \left \| \widehat {\mathbf{y}} \right \|_{\mathbf{W}^{(r)}})\right]
    \\
    & =
    \log
    	\sum_{\widehat {\mathbf{y}}\in\mathcal{Y}(1,N)}\exp(\alpha \left \| \widehat {\mathbf{y}} \right \|_{\mathbf{W}^{(r)}})
    \\
    & =
    \mu_N^{(r)}.
\end{align}

by the definition \eqref{eqn:mudefinition}.

The first two terms inside the final expectation mentioned above (on line \eqref{eqn:firsttwo}) may also be efficiently re-factored; \textit{e.g.} for the second term,

\begin{align*} 
    & \theexpectation\left[\left \| \mathbf{y} \right \|_{\mathbf W^{(r)}}\right]
    \\
    & =
    \sum_{\mathbf{y} \in \mathcal{Y}(1,N)}
    \mathcal{D}(\mathbf{y} | \mathcal{R}, \mathbf{W}, \alpha)
    \sum_{(u,v)\in\mathbf{y}} w^{(r)}_{u,v}
    \\
    & =
    \sum_{\mathbf{y} \in \mathcal{Y}(1,N)}
    \prod_{(u,v)\in \mathbf{y}} \pi_{u,v}
    \sum_{(u,v)\in \mathbf{y}} w^{(r)}_{u,v}
    \\
    & =
    \sum_{(u,v) \in \mathcal{E}} w^{(r)}_{u,v}
    \underbrace{\sum_{\{\mathbf{y} \in\mathcal{Y}(1,N): (u,v)\in \mathbf{y} \}}\prod_{(u',v')\in\mathbf{y}} \pi_{u',v'}}_{\equiv \omega_{u,v}}.
\end{align*}

The expectation of \eqref{eqn:firsttwo}-\eqref{eqn:secondtwo} may therefore be rewritten as \eqref{eqn:kld}.
	\qed
}

\section{Gumbel Softmax Trick}\label{sec:gumbelsoftmaxtrick}

In this section, we investigate a possible interpretation of achieving reparameterization trick (i.e., Gumbel softmax trick) on BDP-VAE framework for interested readers. Compared to the log-derivative trick discussed in \cref{sec:learning}, Gumbel softmax trick samples soft latent optimal paths and has advantages on gradient-based optimization. However, this method may require extra effort in designing its implementation and temperature parameter $\tau$.

\subsection{Node-wise Gumbel Softmax Propagation}

% \textit{--- this section is grayed out because we can hopefully drop it and use only the more elegant log-derivative method described in \autoref{sec:dtw:training}, possibly returning to the reparameterisations in a later project ---}

Recall that the \textit{Gumbel argmax} reparameterisation for the categorical distribution employs parameter-independent Gumbels via 
\begin{align}
	\forall i \in \{1,2,\dots,m\}, ~~ G_i & \sim \mathcal G(0) \\
	k & = \argmax_{i \in \{1,2,\dots,m\}} \log \mu_i + G_i,
	\label{eqn:gumbelargmax:restated}
\end{align}
which is equivalent to 
\begin{align}
	k\sim\mathrm{Categorical}(\beta_1, \beta_2, \dots, \beta_m),
\end{align}
where the probability $\beta_k$ is equal to the r.h.s. of \eqref{eqn:gumbelargmax}.

The \textit{Gumbel softmax} is a differentiable approximation to the above, where the $k\in\{1,2,\dots,m\}$ of \eqref{eqn:gumbelargmax:restated} is replaced by $\mathbf{\kappa} \in\mathbb R^{m}$ given by
\begin{align}
	\kappa_k = \frac{\exp\big((\log \mu_k + G_k)/\tau\big)}{\sum_{i=1}^m\exp\big((\log \mu_i + G_i)/\tau\big)},
\end{align}
where $\tau$ is a free parameter. This approximation is useful for gradient-based optimization because it is both differentiable and parameterised.

Due to the generally intractable size $\left|\mathcal Y(1,N)\right|$ of the set of paths that make up the domain of $\mathcal D(\mathbf{y} | \mathcal R, \mathbf{W}, \alpha)$, there is no simple analogue of the above for our setting. Instead, we offer two alternative approaches, both of which are differentiable reparameterizations of a distribution of real values, one per node.

\subsubsection{Marginal Node-wise Gumbel Softmax Distribution}
\label{sec:marginalsoftmax}

Consider the following
\begin{definition}
Let $\mathbf{y} \sim \mathcal D(\mathcal R, \mathbf{W}, \alpha)$. The hitting probability for the node $v$ is the probability that $\mathbf{y}$ includes v,
\begin{align}
	\zeta_v \equiv p(v\in \mathbf{y}).
	\label{eqn:hitting}
\end{align}
\end{definition}
The node-hitting probabilities may be efficiently obtained via 
\begin{lemma}
	\label{lem:zetarecursion}
	The $\zeta_u$ obey the recursion
	\begin{align}
		\zeta_N & = 1 \\
		\zeta_u & = \sum_{v\in\mathcal C(u)} \zeta_v \, \pi_{u,v},
	\end{align}
	for all $v\in\left\{N-1,N-2,\dots,1\right\}$ (in reverse topological order w.r.t. $\mathcal R$).
\end{lemma}
We may now simply associate with each node $u\in\mathcal V$ a Bernoulli random variable with probability parameter $\zeta_u$, along with a Gumbel-softmax approximation of that Bernoulli --- see \autoref{sec:binarygumbelmax} for details.

\subsubsection{Path-dependent Node-wise Gumbel Softmax Distribution}
\label{sec:pathdependentsoftmax}

Alternatively, we can soften the path sampling algorithm of \cref{cor:sample} to obtain a distribution over $\{\gamma_u\in [0,1] \}_{u\in\mathcal V}$ that better captures the dependence between the events $u\in \mathbf y$ for all $u\in\mathcal V$. That is, we let
\begin{align}
	\gamma_N & = 1 \\
	\gamma_u & = \sum_{v\in\mathcal C(u)} \gamma_v \, \delta_{u,v},
\end{align}
for all $u\in\left\{N-1,N-2,\dots,1\right\}$ (in reverse topological order w.r.t. $\mathcal R$), where $\delta_{u,v}$ is the Gumbel-softmax analog to step 2 of \cref{cor:sample}, namely
\begin{align}
\forall (u,v) \in \mathcal E, ~ G_{u,v} & \sim \mathcal G(0)
\\
	\delta_{u,v} & = 
	\frac{\exp\big((\log \pi_{u,v} + G_{u,v})/\tau\big)}{\sum_{i\in\mathcal C(u)}\exp\big((\log \pi_{u,i} + G_{u,i})/\tau\big)}.
\end{align}

\subsection{KL Divergence Between two Gumbels}

\begin{lemma}
	The KL divergence between unit variance Gumbels is
\begin{align}
	\kl{\mathcal{G}(\alpha)}{\mathcal{G}(\beta)}
	& = 
	\alpha-\beta + (1-\exp(\alpha-\beta))\ei(-\exp(-\alpha)),
\end{align}
in terms of the standard special ``exponential integral'' function
\begin{align}
	\ei(z) \equiv -\int_{-z}^\infty\exp(-t)/t \mathop{}\!\mathrm{d} t.
\end{align}
\end{lemma}

\proof{
\begin{align*}
	&
	\kl{\mathcal G(\alpha)}{\mathcal G(\beta)}
	\\
	& = 
	\int_0^\infty G(x|\alpha)\log \frac{G(x|\alpha)}{G(x|\beta)} \mathop{}\!\mathrm{d} x
	\\
	& = 
	\int_0^\infty \exp(-(x-\alpha)-\exp(x-\alpha)) \log \frac{\exp(-(x-\alpha)-\exp(x-\alpha))}{\exp(-(x-\beta)-\exp(x-\beta))} \mathop{}\!\mathrm{d} x	
	\\
	& = 
	\int_0^\infty \exp(-(x-\alpha)-\exp(x-\alpha)) (\alpha-\beta+\exp(x-\beta)-\exp(x-\alpha))\mathop{}\!\mathrm{d} x	
	\\
	& = 
	\alpha-\beta + I_1 - I_2
	\\
	& = 
	\alpha-\beta + (1-\exp(\alpha-\beta))\ei(-\exp(-\alpha)),
\end{align*}
because
\begin{align*}
	I_2 & \equiv 
	\int_0^\infty \exp(-\exp(x-\alpha))\mathop{}\!\mathrm{d} x	
	\\
	& 
= 
	-\ei(-\exp(-\alpha))
\intertext{and}
	I_1 & \equiv 
	\int_0^\infty \exp(\alpha-\beta-\exp(x-\alpha))\mathop{}\!\mathrm{d} x	
	\\
	& =
 -\exp(\alpha-\beta)\ei(-\exp(-\alpha)),
\end{align*}
giving the desired result. \qed
}
\subsection{Binary Gumbel (Soft) Max}

\label{sec:binarygumbelmax}

For the binary case, we can sample just one logistic random variable, rather than two Gumbels, as per the traditional Gumbel max trick. Let $X\sim\mathrm{Bernoulli}(\zeta)$. The Gumbel max parameterised, for $g_1, g_2 \sim \mathcal G(0)$
\begin{align}
	X 
	& = 
	\argmax_{k\in \{1,2\}} \left(\log \zeta + g_1, \log (1-\zeta) + g_2\right)_k
	\\
	& = 
	\argmax_{k\in \{1,2\}} \left(\log \zeta, \log (1-\zeta) + l\right)_gbk,
\end{align}
where we may show that $l=g_2-g_1\sim \mathrm{Logistic}(0,1)$.\footnote{\url{https://en.wikipedia.org/wiki/Logistic_distribution}} 
The soft-max analogue $X_\tau\in(0,1)$ of $X$, where $\tau$ is a temperature parameter, is therefore
\newcommand\tmo{\tau^{-1}}
\begin{align}
	X_\tau 
	& \equiv \frac{\exp(\tmo \log \zeta)}{\exp(\tmo \log \zeta)+\exp\Big(\tmo \big(\log (1-\zeta)+l\big)\Big)}.
\end{align}

\begin{figure}[ht]
\begin{center}
\centerline{\includegraphics[width=0.6\columnwidth]{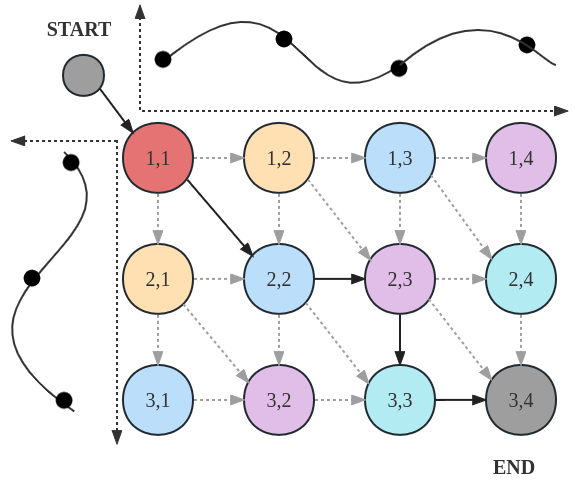}}
\caption{Computational graph $\mathcal{R}$ of the DTW algorithm}
\label{fig.DTW-DAG}
\end{center}
\vskip -0.2in
\end{figure}

\section{Examples of computational graphs} \label{D:example}
We demonstrate two examples of how to implement the \textit{Bayesian dynamic programming} to obtain structural latent optimal paths on different structured computational graphs.

\subsection{Dynamic Time Warping} \label{sec:DTW}

We first extend the Bayesian dynamic programming to the dynamic time warping (DTW) algorithm~\citep{sakoe1978dynamic, mensch2018differentiable} which aims to seek an optimal alignment path with the maximum score (or minimum cost) given two time series. Given two time-series $\mathbf{A}$ and $\mathbf{B}$ with lengths $N_A$ and $N_B$. Let $a_j$ and $b_i$ be $j^{th}$ and $i^{th}$ observations of $\mathbf{A}$ and $\mathbf{B}$, respectively. We denote $\mathbf{W} \in \mathbb{R}^{N_B \times N_A}$ as a pair-wise weight matrix,
% similarity matrix 
% $\mathbf{W} \in \mathbb{R}^{N_B \times N_A}$,
% as the weight matrix, 
where $w_{ij}$ represents similarity measurement between observation point $b_i$ and observation point $a_j$, namely, $w_{ij} = d(b_i,a_j)$, where function $d(\cdot)$ is an arbitrary similarity metric. The $\mathcal{Y}$ defines the set of the population of all possible time-series alignment paths $\mathbf{y}$, in which the path connects the upper-left $(1,1)$ node to the lower-right $(N_B, N_A)$ node with $\rightarrow$, $\downarrow$ and  $\searrow$ moves only.

Following \cref{lem:murecursion}, we can obtain the location parameters $\mu \in \mathbb{R}^{N_B \times N_A}$ given the DAG $\mathcal{R}$ and $\mathbf{W}$.
% the location parameters $\mu \in \mathbb{R}^{N_B \times N_A}$ can be interpreted as a cumulative weight matrix given $\mathcal{R}$ and $\mathbf{W}$.
Then the optimal path $\mathbf{y}$ can be sampled reversely according to the transition matrix $\pi \in \mathbb{R}^{N_B \times N_A \times 3}$. The probability of transition $(v,v') \rightarrow (u,u')$, where $(u,u') \in \mathcal{P}(v,v')$, is defined as

\begin{equation}\label{eqn:DTW_pi}
    \pi_{(u,u'),(v,v')} \equiv p(y_{i-1} = (u,u')| y_i = (v,v'),  (u,u') \in \mathcal{P}(v,v')) 
\end{equation}
for all $i\in \{ (1,1),..,(N_B,N_A)\}$

\cref{fig.DTW-DAG} is an example of the computational graph of DTW with $N_A =4$ and $N_B = 3$. The bold black arrows indicate one aligned path $\mathbf{y} \in \mathcal{Y}$ on the DAG. Pseudocode to compute the location parameter $\mu$ and transition matrix $\pi$ and sampling optimal path $\mathbf{y}$ are provided in \cref{alg:DTW_forward} and \cref{alg:DTW_backward}.

\begin{algorithm}[tb]
   \caption{Compute $\mu$ and $\pi$ on DTW}
   \label{alg:DTW_forward}
\begin{algorithmic}
   \STATE {\bfseries Input:} Weight matrix $\mathbf W \in \mathbb{R}^{N_B \times N_A}$, $\alpha$;
   \STATE Initialize $\mu \in \mathbb{R}^{(N_B +1)\times (N_A+1)}$,
   \STATE  $\mu_{i,0} = -\infty$, $\mu_{0,j} = -\infty$, $i\in [ N_B], j \in [N_A]$;
   \STATE  $\mu_{0,0} = 0$;

   \FOR{$i = 1$ {\bfseries to} $N_B$, $j = 1$ {\bfseries to} $N_A$ }
        \STATE $\mu_{i,j} = \log(\exp(\mu_{i-1,j-1}+\alpha w_{i,j})+\exp(\mu_{i,j-1} +\alpha w_{i,j})+ \exp(\mu_{i-1,j} +\alpha w_{i,j}))$
   \ENDFOR
   \STATE Initialize $\pi_{i,j} = \mathbf{0}_3$, $i \in [N_B], j \in [N_A]$; 
   \FOR{$i = N_B$ {\bfseries to} $1$ ,$j = N_A$ {\bfseries to} $1$}
            \STATE $\pi_{i,j} = [\frac{\exp(\mu_{i,j-1}+\alpha w_{i,j})}{\exp(\mu_{i,j})},
            \frac{\exp(\mu_{i-1,j-1}+\alpha w_{i,j})}{\exp(\mu_{i,j})}, \frac{\exp(\mu_{i-1,j}+\alpha w_{i,j})}{\exp(\mu_{i,j})}]$
   \ENDFOR
\end{algorithmic}
\end{algorithm}

\begin{algorithm}[tb]
   \caption{Sample stochastic optimal path $y$ on DTW}
   \label{alg:DTW_backward}
\begin{algorithmic}
   \STATE {\bfseries Input:} Transition matrix $\pi \in \mathbb{R}^{N_B \times N_A \times 3}$
   \STATE Initialize $y \in \mathbb{R}^{N_B \times N_A}$, $i = N_B, j =N_A$; 
   \STATE $y_{N_B,N_A} =1$, $y_{/\ N_B,/\ N_A} = 0$;

    \WHILE{$i>0$ and $j > 0$}
        \STATE x $\sim \text{Categorical}(\pi_{i,j})$
        \IF{x = 0}
            \STATE $y_{i,j-1} = 1$
            \STATE $ j = j-1$
        \ENDIF
        \IF{x = 1}
            \STATE $y_{i-1,j-1} = 1$
            \STATE $i= i-1$
            \STATE $j=j-1$
        \ENDIF
        \IF{x = 2}
            \STATE $y_{i-1,j} = 1$
            \STATE $i= i-1$
        \ENDIF
    \ENDWHILE
\end{algorithmic}
\end{algorithm}

We denote the marginal probability of edges between node $(u,u')$ and node $(v,v')$, where $(u,u') \in \mathcal{P}(v,v)$, as $\omega_{(u,u')(v,v')}$. Following~\cref{lem:omega}, $\omega$ can be computed by \cref{eqn:omega_dtw}

\begin{equation}\label{eqn:omega_dtw}
% \begin{split}
%     \omega_{i,j} =  & [\lambda_{i,j-1}\pi_{(i,j),(i,j-1)}\rho_{i,j},\lambda_{i-1,j-1}\pi_{(i,j),(i-1,j-1)}\rho_{i,j}, \lambda_{i-1,j}\pi_{(i,j),(i-1,j)}\rho_{i,j}]
% \end{split}
\omega_{(u,u'),(v,v')} = \pi_{(u,u'),(v,v')}\lambda_{(u,u')}\rho_{(v,v')}
\end{equation}

For implementation, the pseudocode to compute the marginal probabilities $\omega$ show on \cref{alg:marginal_dtw}.

\begin{algorithm}[ht]
   \caption{Compute marginal probability of edges $\omega$ on DTW}
   \label{alg:marginal_dtw}
\begin{algorithmic}
   \STATE {\bfseries Input:} Transition matrix $\pi \in \mathbb{R}^{N_B \times N_A \times 3}$
   \STATE Initialize $\omega \in \mathbb{R}^{N_B \times N_A \times 3}$; $\omega_{i,j} = 0$; $i \in [N_B], j\in [N_A]$
   \STATE $\lambda,\rho \in \mathbb{R}^{(N_B+1) \times (N_A+1)}$; 
   \STATE $\lambda_{i,j} = 0$; $i\in [N_B], j\in [N_A]$; $\lambda_{0,0} = 1$;
   \STATE $\rho_{i,j} = 0$, $i\in [N_B], j\in [N_A]$; $\rho_{N_B,N_A} = 1$;
   
   \COMMENT{\textcolor{gray}{Topological iteration for $\lambda$}}
    \FOR{$i=1$ {\bfseries to} $N_B$, $j=1$  {\bfseries to}  $N_A$}
            \STATE $\lambda_{i,j} = [\lambda_{i,j-1},\lambda_{i-1,j-1}, \lambda_{i-1,j}]\pi_{i,j}^T$
    \ENDFOR
    
    \COMMENT{\textcolor{gray}{Reversed iteration for $\rho$}}
    \FOR{$i=N_B$ {\bfseries to} $1$, $j=N_A$ {\bfseries to} $1$}
            \STATE $\rho_{i,j} = [\rho_{i,j+1},\rho_{i+1,j+1},\rho_{i+1,j}][\pi_{i,j+1,0},\pi_{i+1,j+1,1},\pi_{i+1,j,2}]^T$
    \ENDFOR

    \COMMENT{\textcolor{gray}{Compute $\omega$}}
    \FOR{$i =1 $ {\bfseries to} $N_B$, $j=1$ {\bfseries to} $N_A$}
            \STATE $\omega_{i,j} = \rho_{i,j}[\lambda_{i,j-1},\lambda_{i-1,j-1},\lambda_{i-1,j}]^T\cdot \pi_{i,j}$
            % \STATE $\omega_{i,j} = [\lambda_{i,j-1}\pi_{(i,j),(i,j-1)}\rho_{i,j},\lambda_{i-1,j-1}\pi_{(i,j),(i-1,j-1)}\rho_{i,j}, \lambda_{i-1,j}\pi_{(i,j),(i-1,j)}\rho_{i,j}]$
    \ENDFOR
\end{algorithmic}
\end{algorithm}

\begin{figure}[ht]
\begin{center}
\centerline{\includegraphics[width=0.6\columnwidth]{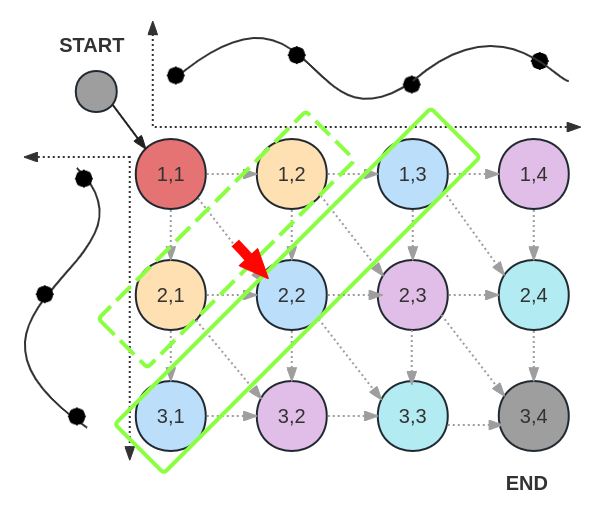}}
\caption{Updating the computational graph of DTW diagonally with linear time complexity to $N_B$ and $N_A$.}
\label{fig.DTW-update}
\end{center}
\vskip -0.2in
\end{figure}

\subsubsection{Time complexity analysis} \label{E:DTW_time}

Following above definition of the computational graph of DTW, where the pair-wise weight matrix $\mathbf{W} \in \mathbb{R}^{N_B \times N_A}$. According to \cref{cor:linear_time}, the edge numbers under DTW graph is $|\mathcal{E}| = 3N_A N_B -2N_A - 2N_B +1$. Therefore, the time complexity in the above pseudo algorithm is $\mathcal{O}(3N_A N_B -2N_A - 2N_B +1)$. Inspired by \citet{tralie2020exact}, we further reduce its time complexity by updating diagonally in parallel as illustrated in \cref{fig.DTW-update}. In this way, the time complexity of computational graph of DTW becomes linear $N_B$ and $N_A$ as $\mathcal{O}(N_A + N_B -1)$.

\subsection{Monotonic Alignment} \label{sec:MA}

\begin{figure}[ht]
\begin{center}
\centerline{\includegraphics[width=0.8\columnwidth]{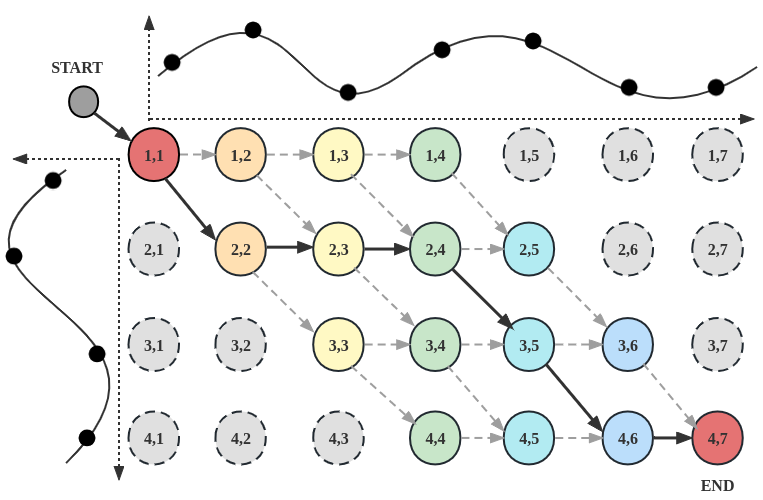}}
\caption{Computational graph $\mathcal{R}$ of the MA algorithm.}
\label{fig.MA-DAG}
\end{center}
\vskip -0.2in
\end{figure}

% Bayesian dynamic programming can also be applied to 
Monotonic alignment (MA) is often used in the field of machine learning, particularly in the context of sequence-to-sequence (Seq2Seq) tasks.
Taking the same definition of two given time-series in \cref{sec:DTW},
$\mathcal{Y}$ defines a population of all possible path sample $\mathbf y$, in which the path connects the upper-left $(1,1)$ node to the lower-right $(N_B,N_A)$ node with $\rightarrow$ and  $\searrow$ moves only, where  $N_B < N_A$.

The location parameters $\mu \in \mathbb{R}^{N_B \times N_A}$ can be computed following \cref{lem:murecursion} and the optimal path $\mathbf{y}$ can be sampled according to the transition matrix $\pi \in \mathbb{R}^{N_B \times N_A \times 2}$. The probability of transition $(v,v') \rightarrow (u,u')$ for $(u,u') \in \mathcal{P}(v,v')$ is defined in \cref{eqn:DTW_pi}.
\cref{fig.MA-DAG} gives an example of the monotonic alignment computational graph with $N_A =7$ and $N_B = 4$. The bold black arrows indicate one possible aligned path.

\begin{algorithm}[ht]
   \caption{Compute $\mu$ and $\pi$ on MA}
   \label{alg:MA_forward}
\begin{algorithmic}
   \STATE {\bfseries Input:} Weight matrix $\mathbf W \in \mathbb{R}^{N_B \times N_A}$, $\alpha$;
   \STATE Initialize $\mu \in \mathbb{R}^{(N_B+1) \times (N_A+1)}$
   \STATE $\mu_{i,j} = -\infty$, $i\in [N_B], j \in [N_A]$;  $\mu_{0,0} = 0$;

   \FOR{$j = 1$ {\bfseries to} $N_A$}
        \FOR{$i=1$ {\bfseries to} min$(j,N_B)$}
            \STATE $\mu_{i,j} = \log(\exp(\mu_{i-1,j-1}+\alpha w_{i,j})+\exp(\mu_{i,j-1} +\alpha w_{i,j}))$
        \ENDFOR
   \ENDFOR
   \STATE Initialize $\pi_{i,j} = \mathbf{0}_2$, $i \in [N_B], j \in [N_A]$; 
   %$\pi_{N_B,N_A} = [0,1]$;
   \FOR{$j = N_A$ {\bfseries to} $1$}
        \FOR{$i = \text{min}(j, N_A)$ {\bfseries to} max$(j-N_A+N_B,1)$}
            \STATE $\pi_{i,j} = [\frac{\exp(\mu_{i,j-1}+\alpha w_{i,j})}{\exp(\mu_{i,j})},\frac{\exp(\mu_{i-1,j-1}+\alpha w_{i,j})}{\exp(\mu_{i,j})}]$
        \ENDFOR
   \ENDFOR
\end{algorithmic}
\end{algorithm}

\begin{figure}[t]
\begin{center}
\centerline{\includegraphics[width=0.8\columnwidth]{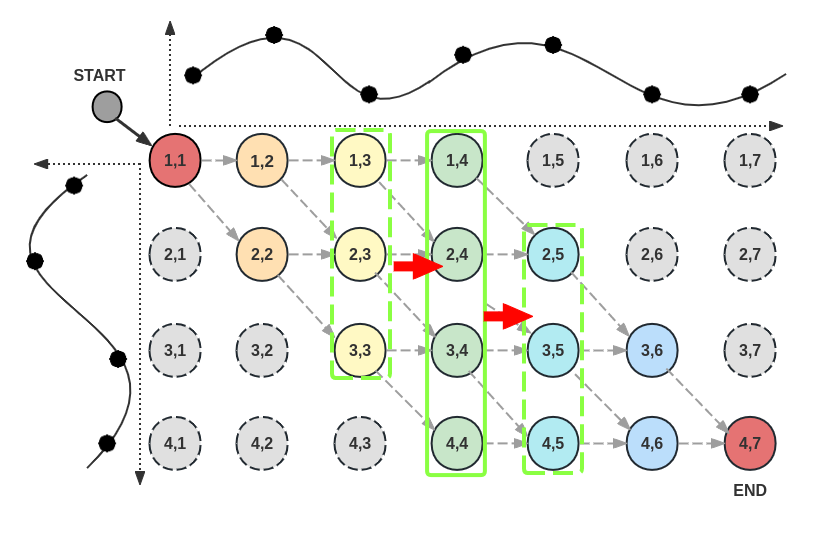}}
\caption{Updating the computational graph of MA vertically with linear time complexity to $N_B$ and $N_A$.}
\label{fig.MA-DAG_update}
\end{center}
\vskip -0.2in
\end{figure}

\begin{algorithm}[t]
   \caption{Sample stochastic optimal path $y$ on MA}
   \label{alg:MA_backward}
\begin{algorithmic}
   \STATE {\bfseries Input:} Transition matrix $\pi \in \mathbb{R}^{N_B \times N_A \times 2}$
   \STATE Initialize $y \in \mathbb{R}^{N_B \times N_A}$, $i = N_B, j =N_A$; 
   \STATE $y_{N_B,N_A} =1$, $y_{/\ N_B,/\ N_A} = 0$;

    \WHILE{$i>0$ and $j > 0$}
        \STATE x $\sim \text{Categorical}(\pi_{i,j})$
        \IF{x = 0}
            \STATE $y_{i,j-1} = 1$
            \STATE $ j = j-1$
        \ENDIF
        \IF{x = 1}
            \STATE $y_{i-1,j-1} = 1$
            \STATE $i= i-1$
            \STATE $j=j-1$
        \ENDIF
    \ENDWHILE

\end{algorithmic}
\end{algorithm}

Pseudo-code for computing the location parameter $\mu$ and transition matrix $\pi$ are provided in \cref{alg:MA_forward}. The sampling algorithm is presented in \cref{alg:MA_backward}.

\begin{algorithm}[t]
   \caption{Compute marginal probability of edges $\omega$ on MA}
   \label{alg:marginal_ma}
\begin{algorithmic}
   \STATE {\bfseries Input:} Transition matrix $\pi \in \mathbb{R}^{N_B \times N_A \times 2}$
   \STATE Initialize $\omega \in \mathbb{R}^{N_B \times N_A \times 2}$; $\omega_{i,j} = 0$; $i \in [N_B], j\in [N_A]$
   \STATE $\lambda,\rho \in \mathbb{R}^{(N_B+1) \times (N_A+1)}$; 
   \STATE $\lambda_{i,j} = 0$; $i\in [N_B], j\in [N_A]$; $\lambda_{0,0} = 1$;
   \STATE $\rho_{i,j} = 0$, $i\in [N_B], j\in [N_A]$; $\rho_{N_B,N_A} = 1$;
   
   \COMMENT{\textcolor{gray}{Topological iteration for $\lambda$}}
    \FOR{$j=1$ {\bfseries to} $N_A$}
        \FOR{$i = 1 $ {\bfseries to} min$(j,N_B)$}
            \STATE $\lambda_{i,j} = [\lambda_{i,j-1},\lambda_{i-1,j-1}]\pi_{i,j}^T$
        \ENDFOR
    \ENDFOR
    
    \COMMENT{\textcolor{gray}{Reversed iteration for $\rho$}}
    \FOR{$j=N_A$ {\bfseries to} $1$}
        \FOR{$i =$ min$(j,N_B)$ {\bfseries to} max$(j-N_A+N_B,1)$}
            \STATE $\rho_{i,j} = [\rho_{i,j+1},\rho_{i+1,j+1}][\pi_{i,j+1,0},\pi_{i+1,j+1,1}]^T$
        \ENDFOR
    \ENDFOR

    \COMMENT{\textcolor{gray}{Compute $\omega$}}
    \FOR{$j=1$ {\bfseries to} $N_A$}
        \FOR{$i = 1 $ {\bfseries to} min$(j,N_B)$}
            \STATE $\omega_{i,j} = \rho_{i,j}[\lambda_{i,j-1},\lambda_{i-1,j-1}]^T\cdot \pi_{i,j}$
            % \STATE $\omega_{i,j} = [\lambda_{i,j-1}\pi_{(i,j),(i,j-1)}\rho_{i,j},\lambda_{i-1,j-1}\pi_{(i,j),(i-1,j-1)}\rho_{i,j}]$
        \ENDFOR
    \ENDFOR
\end{algorithmic}
\end{algorithm}

We then denote the marginal probability of edges between node $(u,u')$ and node $(v,v')$, where $(u,u') \in \mathcal{P}(v,v')$, as $\omega_{(u,u'),(v,v')}$. $\omega$ can be computed by \cref{eqn:omega_dtw} and the pseudocode to compute the marginal probabilities $\omega$ show on \cref{alg:marginal_ma}.

\subsubsection{Time complexity analysis} \label{E:MA_time}

Following the setting of computational graph of MA in above, where the pair-wise weight matrix $\mathbf{W}\in \mathbb{R}^{N_B \times N_A}$ and the graph is defined in \cref{fig.MA-DAG}, where $N_B < N_A$. The edge number under MA graph is $|\mathcal{E}| = 2N_B N_A -2N_B^2 - N_A +2 N_B -1$ and the time complexity in pseudo algorithms of computational graph of MA is $\mathcal{O}(2N_B N_A -2N_B^2 - N_A +2 N_B -1)$. By updating the graph vertically as illustrated in \cref{fig.MA-DAG_update}, the time complexity can be further reduced to $\mathcal{O}(N_A -1)$.

\section{Details of model architecture in experiments} \label{E:model}

\begin{figure}[ht]
\vskip 0.2in
\begin{center}
\centerline{\includegraphics[width=0.8\columnwidth]{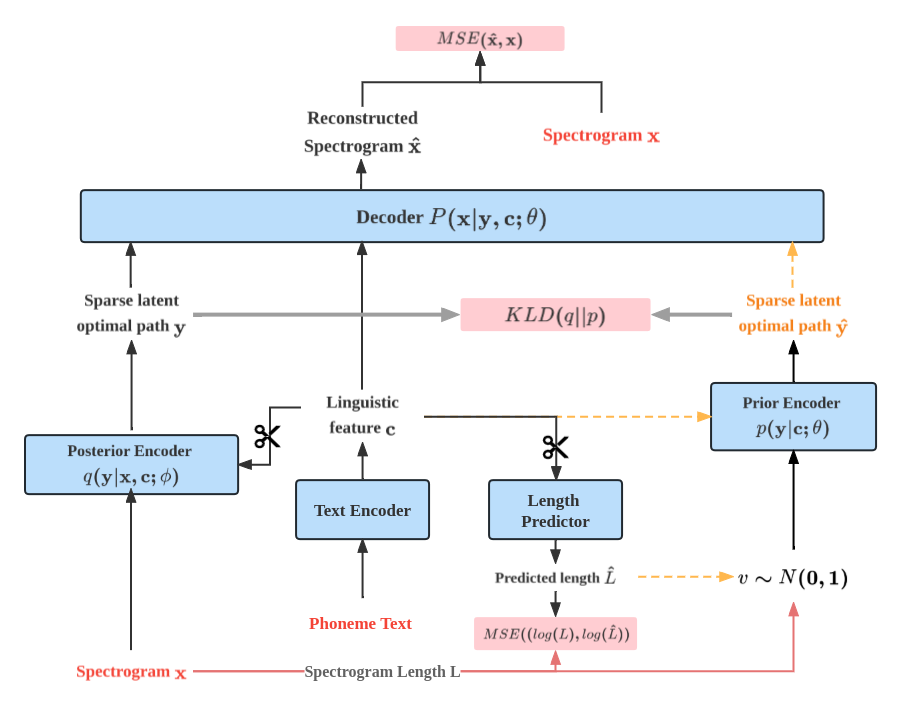}}

\caption{Architecture of BDP-VAE on experiments. The red lines are only turned on during training. Black lines stand for the training process, while yellow dotted lines stand for the inference phase. The scissors represent the gradient not being back-propagated along with the arrow.}
\label{fig.VAE_TTS}
\end{center}
\vskip -0.2in
\end{figure}

Given a spectrogram input $\mathbf x = [x_1,...,x_t] $ and a corresponding phoneme text $\mathbf c' = [c'_1,...,c'_n]$ , where $t$ and $n$ are the lengths of the input sequences. We assume that there is an unobserved structural sparse optimal path (i.e., monotonic alignment) representation $\mathbf y \in \mathbb{R}^{t \times n}$ aligns $\mathbf x$ and $\mathbf c$ by $\{0,1\}$ under defined DAG $\mathcal{R}$. 

The proposed overall architecture is shown in \cref{fig.VAE_TTS} and the conditional ELBO is
\begin{equation}
\begin{split}
	\mathcal L(\phi, \theta, \mathbf x | \mathbf c)
	= 
	\mathbb E_{\mathbf y \sim q(\cdot|\mathbf x, \mathbf c; \phi)} 
	\left[
		\log p(\mathbf x|\mathbf y, \mathbf c; \theta)
	\right] \\
	- 
	\kl{q(\mathbf y|\mathbf x, \mathbf c; \phi)}{p(\mathbf y|\mathbf c; \theta)}.
\end{split}
% \label{eqn:elbo_vae}
\end{equation}

\begin{figure}[ht]
\centering
\begin{minipage}{.45\textwidth}
  \centering
  \includegraphics[width=.8\linewidth]{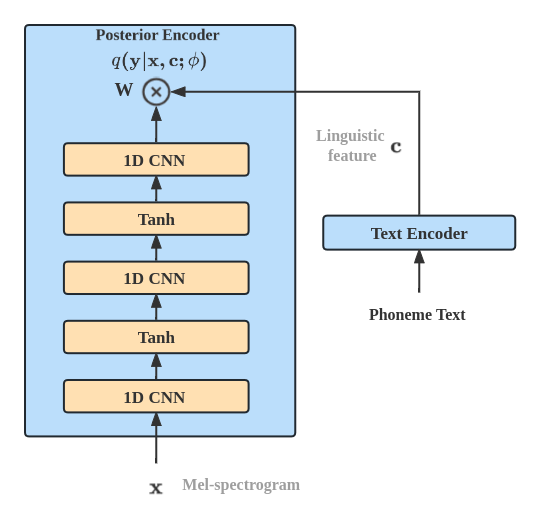}
  \captionof{figure}{The posterior encoder architecture, where the $\bigotimes$ represents \cref{eqn:d}.}
  \label{fig:posterior}
\end{minipage}%
\hfill
\begin{minipage}{.45\textwidth}
  \centering
  \includegraphics[width=.6\linewidth]{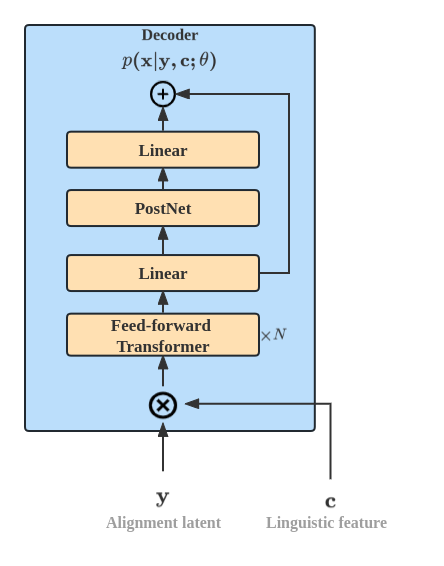}
  \captionof{figure}{The decoder architecture, where the $\bigotimes$ represents the matrix multiplication.}
  \label{fig:decoder}
\end{minipage}
\vfill
\begin{minipage}{.5\textwidth}
  \centering
  \includegraphics[width=\linewidth]{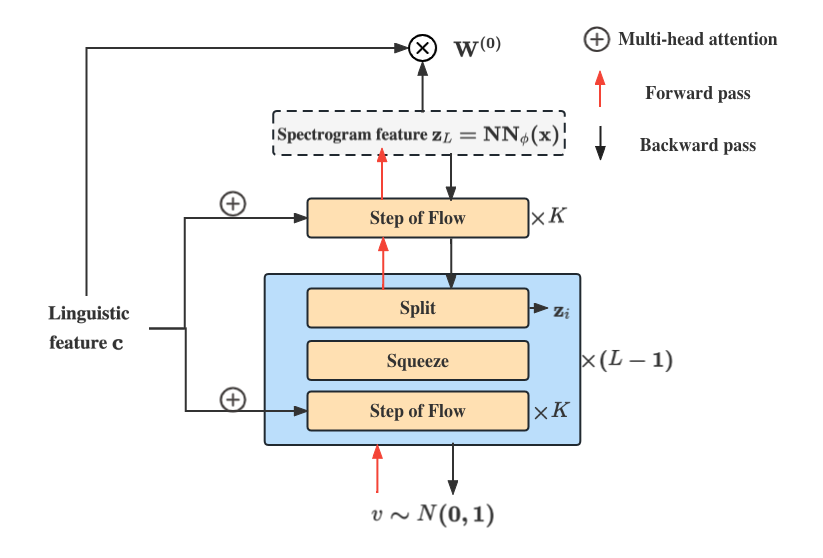}
  \captionof{figure}{The conditional prior encoder architecture, where the $\bigotimes$ represents \cref{eqn:d}.}
  \label{fig:prior}
\end{minipage}
\end{figure}

\subparagraph{Text Encoder}
The text encoder is used to extract a higher level of linguistic features $\mathbf c \in \mathbb{R}^{n\times 256}$ from phoneme text, which adopts the same structures as the one in FastSpeech2 \citep{ren2020fastspeech}, which contains 4 feed-forward transformer (FFT) blocks with 2 multi-head attentions.

\subparagraph{Posterior Encoder} The posterior encoder $q(\mathbf y| \mathbf x, \mathbf c; \phi) = \mathcal{D}(\mathbf y| \mathcal{R}, \mathbf{W} = d(\text{NN}_\phi(\mathbf x),\mathbf c), \alpha)$, where $\alpha$ is a preset hyper-parameter. Note that the gradient with respect to $\theta$ will not backpropagate to the text encoder. The architecture of the posterior encoder shouldn't be too complex, its goal is to extract temporal information from spectrogram inputs $\mathbf{x}$ by a neural network model parameterized by $\phi$. In the posterior encoder (\cref{fig:posterior}), the spectrogram is fed into a convolution-based PostNet \citep{postnet} to upsample the feature dimension to the size of the linguistic feature $\mathbf c$, i.e. $\text{NN}_\phi(\mathbf{x}) \in \mathbb{R}^{t\times 256}$. The final weight matrix $\mathbf W \in \mathbb{R}^{t\times n}$ computed by

\begin{equation} \label{eqn:d}
    d(f_\phi(\mathbf x), \mathbf c) =  softmax( \text{NN}_\phi(\mathbf x) \mathbf c^T)
\end{equation}
where the softmax function is applied over the $t$ dimension.

\subparagraph{Decoder} The architecture of the decoder is also the same as the one in FastSpeech2 \citep{ren2020fastspeech}. In the decoder (\cref{fig:decoder}), the latent optimal path $\mathbf y$ and the linguistic feature $\mathbf c$ are extended by a matrix multiplication which extends the linguistic feature from length $n$ to length $t$ according to the information of sampled latent optimal path $\mathbf y$, then followed by 4 Feed-forward Transformer blocks with 2 multi-head attentions. We add a residual connection after the linear layer with a PostNet to get the final output.

\subparagraph{Prior Encoder} The prior encoder $p(\mathbf{y}|\mathbf{c};\theta) = \mathcal{D}(\mathbf{y}|\mathcal{R},\mathbf{W}^{(0)} = d(f_\theta(\mathbf{c}),\mathbf{c}),\alpha)$. Following \citet{lu2021vaenar,flowseq}, we make use of a Glow \citep{kingma2018glow} structure to infer spectrogram features conditioned on linguistic features. It consists of multiple Glow blocks. Each of the blocks has an actnorm layer, an invertible $1\times 1$ convolutional layer, and an affine-coupling layer. The transformation network in the affine-coupling layer is based on the Transformer decoder, which the spectrogram feature $\text{NN}_\phi(\mathbf{x})$ as the query and the linguistic feature $\mathbf{c}$ as the key and value. During training, we use the backward pass to infer the probability of the KL divergence. The forward pass is used to generate spectrogram features from the condition and then form the weight matrix to sample latent $\mathbf{\hat y}$ during inference. Details of the architecture are in \cref{fig:prior}.

\subparagraph{Length Predictor} Following \citet{lu2021vaenar}, the length predictor consists of a 1-channel fully connected layer with ReLU activation. The length predictor is optimized by an MSE loss, and the gradient will not propagate to the text encoder.

\section{Experimental details} \label{EE:exp}

\subsection{Experimental Setup}
% \noindent\textbf{Experimental Setup}
All experiments were performed on one NVIDIA GeForce RTX 3090. To reduce the variance of gradients during training, we applied the variance reduction technique with a 5-iteration moving average baseline on the REINFORCE loss. The moving average baseline technique here is used for reducing the gradient variance for the REINFORCE estimator.

\subsection{Experimental Details for End-to-end Text-to-speech} \label{G:details}

In the experiment of the end-to-end text-to-speech on the computational graph of MA in \cref{sec:MA}, the latent space captures the discrete monotonic optimal path between phoneme tokens and spectrogram frames. We down-sampled the audio waveform files from 44.1kHz to 22.05kHz and extracted the Mel-spectrograms with 1024 frame size, $25\%$ overlapping, and 80 Mel-filter bins. The model is trained by 18 batch size with a learning rate of 1.25$\text{e}^{-4}$, temperature parameter $\alpha$ of 5 for 700k steps. 

In the evaluation part, we obtain the DTW-MCD score by the package of \textit{pymcd.mcd}\footnote{\url{https://github.com/chenqi008/V2C}}. The 60-iteration Grinffin-Lim Algorithm approximates all the synthesized waveforms. We trained all the models in Table 1 with the same pre-processing setting, and the details are: 

\textbf{FastSpeech2} FastSpeech2~\citep{ren2020fastspeech} is a non-end-to-end TTS model with additional inputs of energy, pitch, phoneme-align TextGrid from MAF\footnote{\url{https://montreal-forced-aligner.readthedocs.io/en/latest/}}. We followed the model configuration the paper provided. We trained the model by 400k iterations and the total loss converged at around $6.9\text{e}^{-3}$.

\textbf{Tacotron2} Tractorn2~\citep{tacotron2} is an end-to-end auto-regressive TTS model that relies on attention to obtain the phoneme duration alignment. We followed the model configuration provided and trained the model by 100k iterations and the total loss converged at around $6.6\text{e}^{-3}$.

\textbf{VAENAR-TTS} VAENAR-TTS~\citep{lu2021vaenar} is an end-to-end utterance-level TTS model that relies on cascade-masked self-attention in the decoder of the VAE to obtain the phoneme duration alignment. We followed the model configuration the paper provided and trained the model by 150k iterations and the total loss converged at around $7.4\text{e}^{-3}$.

\textbf{Glow-TTS} Glow-TTS~\citep{kim2020glow} is an end-to-end phoneme-level TTS model that relies on the monotonic alignment search to obtain the hard phoneme duration alignment. We followed the model configuration the paper provided and trained the model by 120k iterations and the total loss converged at around $-2.1$.

% \textbf{BVAE-TTS} BVAE-TTS~\cite{lee2021bidirectional}

\subsection{Experimental Details for the End-to-end Singing Voice Synthesis}
In the experiment of the end-to-end singing voice synthesis on the computational graph of MA in \cref{sec:MA}, we extract the Mel-spectrogram by the same pre-processing setting as the experiment of end-to-end TTS. The model is trained by the same architecture configuration with 22 batch sizes and a learning rate of $1.25\text{e}^{-4}$ for 200k steps. The 60-iteration Grinffin-Lim Algorithm approximates all the synthesized waveforms.
% We use the same pre-processing and training configuration in~\cref{sec:tts}. 

\subsection{Experimental Details for Latent optimal path on the computational graph of DTW}

In the experiment of the latent optimal path on the computational graph of DTW in \cref{sec:DTW}, we use the TIMIT dataset~\citep{SWVENO_1993} which is recorded by a 16kHz sampling rate. Therefore, we extract the Mel-spectrogram with 1024 frame size, $25\%$ overlapping, and 80 filter channels under a 16kHz sampling rate. The model is trained with batch size 48 and learning rate 2.5$\text{e}^{-4}$.

\begin{figure*}[t]
\vskip 0.2in
\begin{center}
\centerline{\includegraphics[width=0.9\textwidth]{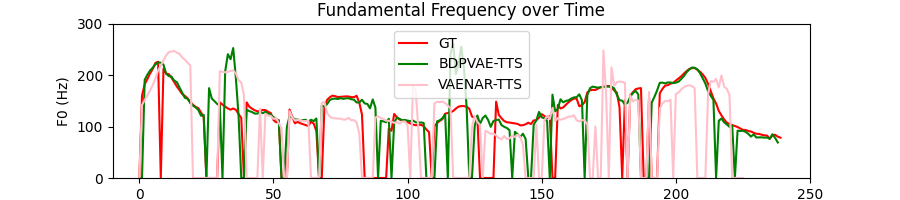}}
\caption{Inference F0 trajectories of utterance "I don't think I can talk about nature without smiling". }
\label{fig.f0_1}

\centerline{\includegraphics[width=0.9\textwidth]{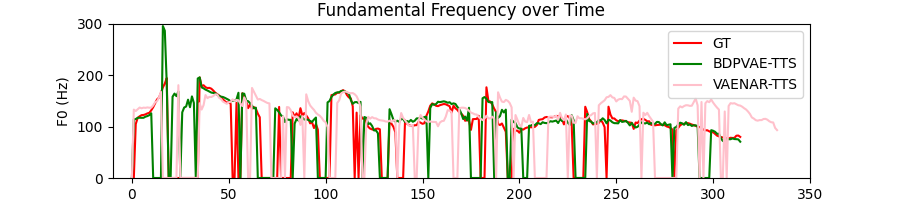}}
\caption{Inference F0 trajectories of utterance "I leaped back into the compartment of the han ship and knelt beside my wilma.". }
\label{fig.f0_2}

\centerline{\includegraphics[width=0.9\textwidth]{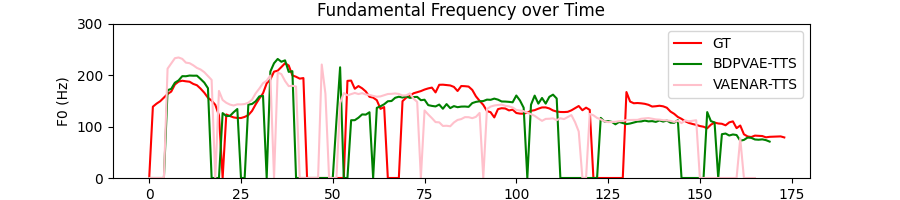}}
\caption{Inference F0 trajectories of utterance "I have reached the end of my explanation". }
\label{fig.f0_3}
\end{center}
\vskip -0.2in
\end{figure*}

\section{Interpretation with the comparison of VAENAR-TTS} \label{sec:F0_compare}

Both BDPVAE-TTS and VAENAR-TTS predict phone-duration alignment on the utterance level. VAENAR-TTS learns the Gaussian latent distribution of learned features of spectrograms and conditions and obtains soft monotonic phoneme-utterance alignment by self-attention with a causality mask in the transformer decoder. Different from VAENAR-TTS, BDPVAE-TTS was adapted from a non-end-to-end phoneme-level TTS model (FastSpeech2~\citep{ren2020fastspeech}) with a Gibbs distribution of stochastic optimal paths $\mathcal{D}(\mathcal{R},\cdot,\alpha)$ defined in \cref{def:gibbs} under monotonic alignment DAG defined in \cref{sec:MA} into a BDP-VAE framework. 
In the inference phase, BDPVAE-TTS reconstructs the spectrogram according to duration-extended phoneme sequences only. 
Since our baseline method (i.e., FastSpeech2) did not outperform the VAENAR-TTS, thus, BDPVAE-TTS also gets a lower performance than VAENAR-TTS.
% We demonstrate more F0 trajectories comparison with the model VAENAR-TTS which gets a lower MCD value than the BDPVAE-TTS. Even though, both models predict phone-duration alignment on the utterance level, the VAENAR-TTS uses cascade attention on the decoder to obtain soft phoneme-utterance monotonic alignment. 

However, soft monotonic alignment in VAENAR-TTS may make the model cannot accurately capture the exact relationship of how phoneme tokens perform in a spectrogram. According to the natural inherent relationship of speech and phonemes, the BDPVAE-TTS obtains sparse monotonic optimal paths in the latent space which are more precise in explaining the relationship between utterance and phoneme tokens. \cref{fig.f0_1} to \cref{fig.f0_3} give additional demonstrations that the synthesized audios from BDPVAE-TTS have closer F0 to the ground truth than VAENAR-TTS. 

\end{document}